# Beyond Prompt Engineering: Neuro-Symbolic-Causal Architecture for Robust Multi-Objective AI Agents

Gokturk Aytug Akarlar
akarlaraytu@gmail.com

*GitHub:* https://github.com/akarlaraytu/Project-Chimera
*Interactive Demo:* https://project-chimera.streamlit.app/

**Abstract**

Large language models show promise as autonomous decision-making agents, yet their deployment in high-stakes domains remains fraught with risk. Without architectural safeguards, LLM agents exhibit catastrophic brittleness: identical capabilities produce wildly different outcomes depending solely on prompt framing. We present Chimera, a neuro-symbolic-causal architecture that integrates three complementary components—an LLM strategist, a formally verified symbolic constraint engine, and a causal inference module for counterfactual reasoning.

We benchmark Chimera against baseline architectures (LLM-only, LLM with symbolic constraints) across 52-week simulations in a realistic e-commerce environment featuring price elasticity, trust dynamics, and seasonal demand. Under organizational biases toward either volume or margin optimization, LLM-only agents fail catastrophically (total loss of \$99K in volume scenarios) or destroy brand trust (−48.6% in margin scenarios). Adding symbolic constraints prevents disasters but achieves only 43-87% of Chimera's profit. Chimera consistently delivers the highest returns (\$1.52M and \$1.96M respectively, some cases +\$2.2M) while improving brand trust (+1.8% and +10.8%, some cases +20.86%), demonstrating prompt-agnostic robustness. Our TLA+ formal verification proves zero constraint violations across all scenarios.

These results establish that architectural design—not prompt engineering—determines the reliability of autonomous agents in production environments. We provide open-source implementations and interactive demonstrations for reproducibility.

## 1 Introduction

The rapid progress of large language models has encouraged their deployment as autonomous decision-making agents in domains such as pricing, retail operations, and supply-chain management. When effective, these systems combine expressive reasoning with scalable optimization, enabling firms to adapt strategic decisions in real time. However, recent studies reveal a fundamental challenge: **LLM-based agents are brittle when used in isolation**, often taking actions that appear rational in local context but violate long-term business logic.

To illustrate, consider a pricing agent whose dual mandate is to grow revenue while preserving brand equity. In our environment simulator, a purely LLM-driven agent frequently overreacts to prompt framing. Under volume-focused instructions, it aggressively reduces prices, triggering repeated catastrophic weeks and a negative annual profit outcome (**–$100K**). When instructed instead to prioritize margins, the same agent overshoots in the opposite direction. Despite achieving short-term gains (**$1.62M** cumulative profit), its decisions erode brand trust from **0.7** → **0.36**, placing future customer lifetime value at risk. These behaviors occur **despite** consistent task definitions and identical state information—highlighting that LLMs alone cannot reliably balance competing business objectives.

Adding a symbolic safeguard improves reliability. An LLM paired with a **Guardian** module ensures actions comply with operational constraints (e.g., avoiding prices below cost). This eliminates catastrophic failures and stabilizes profit outcomes to the **$0.65M–$1.70M** range, depending on strategic bias. Yet this architecture remains **reactive**: it prevents harmful decisions but cannot anticipate trade-offs or evaluate long-term causal consequences of pricing and demand dynamics.

We introduce **Chimera**, a neuro–symbolic–causal architecture designed to combine flexible reasoning, formal safety guarantees, and counterfactual foresight. The neural component (GPT-4) proposes strategic hypotheses. A symbolic verifier—specified and model-checked using **TLA+** ensures each action satisfies economic invariants (e.g., price floors, minimum margin constraints). Meanwhile, a causal inference model learns the firm-specific structural relationships between price, demand, and brand trust, enabling forecasts of **"what would happen if"** scenarios rather than merely predicting short-term reward.

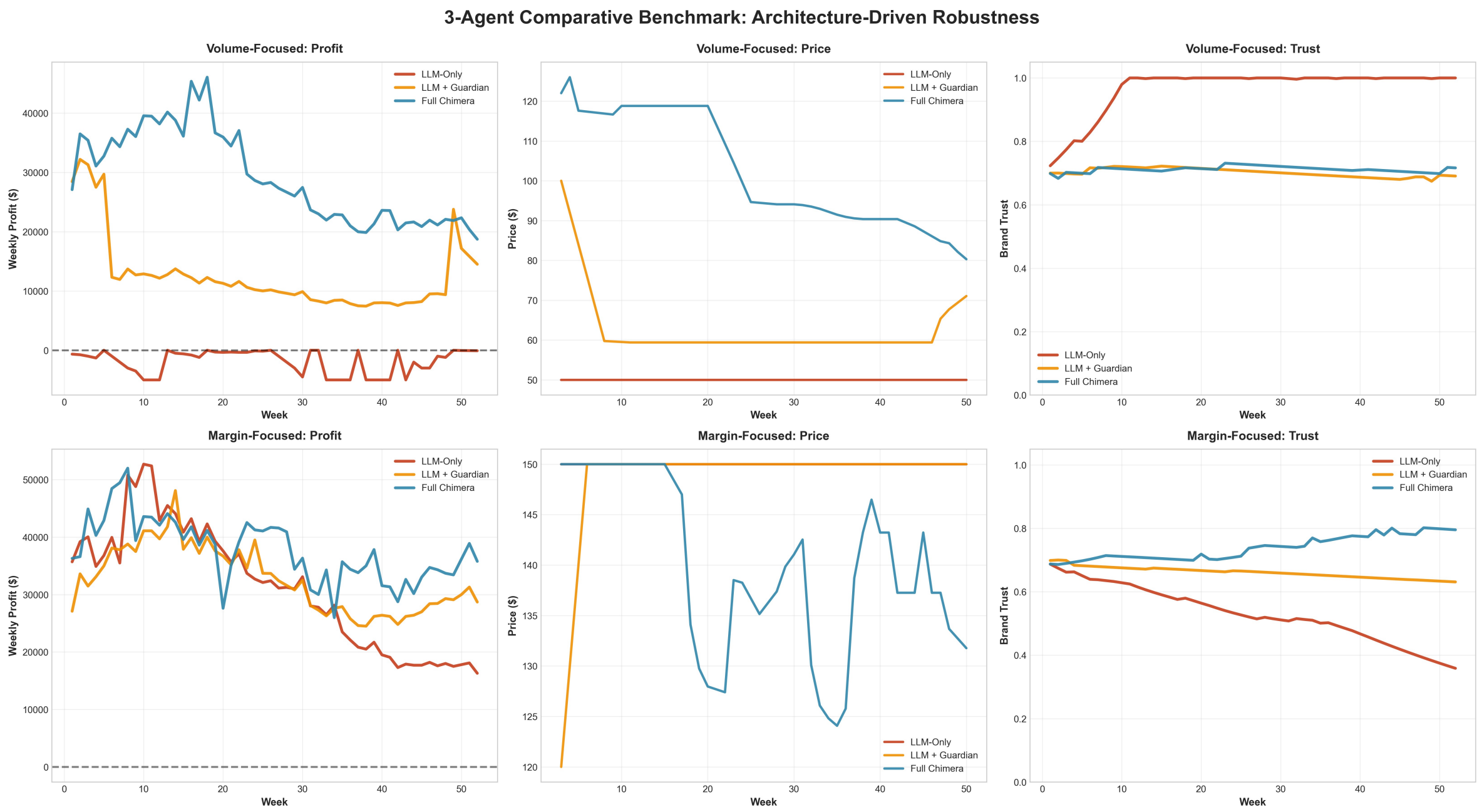

**Figure 1. LLM-Only Agent Failures Under Organizational Biases.** A Panel shows volume-focused prompt leading to catastrophic losses (−$99K total profit, 82.7% failure rate). D shows margin-focused prompt achieving short-term gains ($1.62M) but destroying brand trust (F Panel −32.8% decline). Same agent architecture produces opposite failure modes depending solely on **prompt framing.**

Empirically, Chimera demonstrates advantages along three key dimensions:

(1) **Performance** – up to **+42% total profit** increase versus the best alternative and **>130%** over LLM-only baselines.

(2) **Stability** – **40–60% lower volatility** in weekly returns and the absence of catastrophic failures across conditions.

(3) **Robustness to Prompt Bias** – consistent performance whether the organization expresses a volume-first or margin-first preference, while simpler architectures degrade under framing changes.

Together, these results indicate that **architectural choices—not prompts alone—determine whether autonomous LLM agents behave like trustworthy strategists or reckless guessers**. Chimera provides a principled path toward production-grade autonomous decision systems that preserve both profitability and customer trust over extended planning horizons.

The contributions of this work are fourfold:

**Architecture:** We present the first neuro-symbolic-causal framework **specifically designed for multi- objective decision-making** under organizational constraints, with open-source implementation and inter- active demonstrations.

**Formal Safety:** We provide ***TLA+*** specifications and verification results proving that our symbolic layer maintains invariants across all possible states, establishing a new standard for provable safety in LLM agent systems.

**Empirical Evidence:** Through controlled benchmarks spanning 156 weeks of simulated market conditions (3 architectures × 2 strategies × 52 weeks), we demonstrate that **architectural design determines agent reliability** more than prompt engineering.

**Reproducibility:** We release comprehensive experimental infrastructure including the e-commerce simulator (EcommerceSimulatorV5), all agent implementations, and Streamlit-based interactive environments where researchers can reproduce and extend our findings.

The remainder of this paper is organized as follows. Section 2 surveys related work in neuro-symbolic AI, causal inference, and LLM agent architectures. Section 3 details the Chimera architecture, including our formal verification approach. Section 4 describes our experimental methodology and the e-commerce simulation environment. Section 5 presents our main results across neutral and biased evaluation scenarios. Section 7 concludes with directions for future work.

## 2 Related Work

The challenge of building reliable autonomous agents sits at the intersection of three research traditions: neuro-symbolic integration, causal inference for decision-making, and LLM-based agent architectures. While each area has produced significant advances independently, their synthesis for robust multi-objective optimization remains underexplored.

### 2.1 Neuro-Symbolic AI

The neuro-symbolic paradigm seeks to combine neural networks' pattern recognition capabilities with symbolic systems' logical reasoning and verifiability [1, 2]. Early work focused on knowledge graph integration, where symbolic structures augment neural models with relational constraints [3]. Systems like Neural Theorem Provers [4] and Logic Tensor Networks [5] demonstrated that hybrid architectures could achieve both learning flexibility and logical consistency. More recent efforts have explored differentiable logic programming [6] and neural module networks [7] that maintain symbolic interpretability while supporting gradient-based learning.

However, these approaches typically operate in settings where symbolic knowledge is *static*—encoded once and queried during inference. Chimera differs fundamentally: **our symbolic component actively *repairs* invalid actions proposed by the neural system, rather than merely rejecting them.** This repair mechanism is critical for autonomous agents, which must produce valid outputs even when facing novel situations that exceed their training distribution. Furthermore, we provide **formal verification** that all repairs preserve safety invariants—a guarantee absent in prior work where symbolic modules are them- selves learned or heuristically defined.

### 2.2 Causal Inference in Decision Systems

Causal reasoning has emerged as a critical component for robust decision-making under distribution shift [8, 9]. Pearl's do-calculus provides a mathematical framework for reasoning about interventions, while modern methods like double machine learning [10] and causal forests [11] enable estimation of heterogeneous treatment effects from observational data. Recent work has applied these techniques to policy learning [12], showing that agents equipped with causal models can generalize better than purely correlational approaches when encountering novel scenarios.

Yet most applications of causal inference in AI remain offline: researchers use causal methods to analyze data or train policies, but the resulting agents do not perform causal reasoning at inference time. Notable exceptions include work on counterfactual policy evaluation [13] and causal reinforcement learning [14], though these typically assume access to environment simulators or require extensive interaction data. Chimera integrates a pre-trained causal model (EconML's CausalForestDML) that the agent queries online to predict counterfactual outcomes before committing to actions. This design enables strategic foresight—*the agent can anticipate that aggressive discounting might boost short-term volume but erode brand trust, reducing long-term profitability*—without requiring online experimentation. The causal module also supports continual learning, updating its predictions as new data accumulates during deployment.

### 2.3 LLM Agent Architectures and Tool Use

The emergence of large language models as general-purpose reasoners has sparked interest in agent architectures where LLMs orchestrate complex tasks [15, 16]. Frameworks like ReAct [17] interleave reasoning and action through prompting, while systems such as AutoGPT and BabyAGI demonstrate autonomous goal pursuit through iterative planning. Tool-augmented LLMs can access calculators, search engines, and code interpreters [18], extending their capabilities beyond pure text generation.

Despite these advances, current LLM agent frameworks face two critical gaps that Chimera addresses. **First, they lack hard safety guarantees.** Existing systems rely on prompt engineering ("be careful," "double-check your work") or soft constraints like constitutional AI [19], but cannot prove that dangerous actions will be prevented. Our formally verified Guardian provides such guarantees. **Second, they lack strategic foresight for multi-objective problems.** An LLM instructed to "maximize profit and maintain trust" has no mechanism to evaluate whether a proposed action actually balances these objectives over relevant time horizons. Without counterfactual reasoning capabilities, the agent cannot distinguish actions that appear profitable in isolation but trigger adverse long-term consequences through trust erosion or market dynamics.

Recent work on tool-using agents has begun to address the latter gap by equipping LLMs with prediction APIs and simulators. Chimera extends this direction by demonstrating that architectural integration of tools—*not merely their availability*—determines agent performance. Our experiments show that even when all agents receive identical instructions to optimize dual objectives, only the architecture with both symbolic validation and causal prediction achieves robust performance (Section 5). This suggests a broader principle: the reliability of LLM agents scales with the sophistication of their non-neural components, not with prompt engineering alone.

## 3 The Chimera Architecture

Chimera's design rests on a deliberate separation of concerns: **the neural component generates strategic alternatives, the symbolic component enforces safety constraints, and the causal component predicts long-term consequences.** This division **is not merely conceptual**—each component is implemented as a distinct module with well-defined interfaces, enabling independent verification, testing, and refinement. The orchestration occurs through explicit tool calls: the LLM does not implicitly "have access" to symbolic or causal reasoning; rather, it must deliberately invoke these capabilities and interpret their outputs to inform its decisions.

The control flow proceeds as follows. At each decision point, the agent observes the current market state (price, brand trust, sales volume, ad spend, seasonality). The LLM generates multiple strategic hypotheses—candidate actions exploring different regions of the decision space (e.g., moderate price reduction with increased advertising, aggressive discounting with minimal ad spend, conservative hold). For each hypothesis, the agent calls `check_business_rules` to validate legality; invalid actions are discarded and replaced. Valid hypotheses then undergo causal evaluation via `estimate_profit_impact`, which returns predicted effects on both profit and brand trust over a multi-week horizon. Finally, the LLM selects the action with the best risk-adjusted profile across both objectives and executes it. This loop repeats weekly, with the causal module periodically retraining on accumulated data to refine its predictions.

### 3.1 The Neural Component: Strategic Orchestration

The neural orchestrator is GPT-4o (temperature=0.9, max_tokens=1000), prompted to generate and evaluate strategic alternatives under dual-objective constraints. Unlike baseline LLM-only agents that receive goal-oriented instructions ("maximize profit through volume"), Chimera's LLM operates under a balanced mandate:

> *Your Strategic Objective: Maximize long-term sustainable profit AND brand trust.*
>
> *Organizational Context: [volume-focused or margin-focused, depending on scenario]*
>
> *Your Unique Advantage: Unlike basic agents, you have two powerful tools:*
>
> 1. `check_business_rules`: Prevents catastrophic decisions
> 2. `estimate_profit_impact`: Predicts long-term profit AND trust outcomes
>
> *Decision Process:*
>
> 1. Phase 1: Generate three diverse strategic hypotheses
> 2. Phase 2: Validate each using `check_business_rules`
> 3. Phase 3: Estimate causal impacts using `estimate_profit_impact`
> 4. Phase 4: Select action optimizing profit-trust balance

This prompt structure enforces a **disciplined reasoning process.** The requirement to generate **three hypotheses prevents premature convergence** on locally optimal actions. The mandatory tool-use work- flow ensures that strategic intuitions are grounded in safety validation and causal prediction. Notably, **the prompt does not specify how to balance profit and trust**—it describes the tools available and demands that the agent use them to make informed trade-offs. This design reflects our architectural philosophy: give the LLM powerful instruments and clear objectives, then let its reasoning capabilities determine the optimal path.

Temperature selection warrants brief discussion. We use 0.9 rather than 0.0 to maintain strategic diversity across weeks—at zero temperature, the agent might converge on **repetitive behavior patterns.** However, this introduces variance in individual decisions. **The key insight is that architectural components (Guardian, Causal Engine) provide stability despite neural stochasticity.** Even when the LLM proposes erratic actions, the Guardian catches unsafe moves and the Causal Engine reveals their long- term consequences, steering decisions toward robust outcomes. This stands in contrast to LLM-only agents, where stochasticity directly propagates to actions, producing the high volatility visible in results.

### 3.2 The Symbolic Component: Formally Verified Safety

The Guardian (`SymbolicGuardianV4`) implements hard constraints derived from business logic and operational realities. Its rule set includes:

- **Price floors:** No selling below cost (price ≥ cost × 1.1 buffer)
- **Price ceilings:** Market tolerance limits (price ≤ $150)
- **Margin requirements:** Ensure sustainable unit economics (margin ≥ 15%)
- **Ad spend caps:** Prevent budget overruns (weekly spend ≤ $5,000)
- **Rate limits:** Bound change velocity (|Δprice| ≤ 40% discount, +50% increase per week; ad increase ≤ $1,000/week)

These constraints are encoded as predicates over state-action pairs. The Guardian exposes two interfaces:

- `validate_action(action, state)` → {is_valid, violations, message} checks whether a proposed action would violate constraints,
- `repair_action(action, state)` → {safe_action, repairs} projects invalid actions onto the nearest valid point in action space.

For example, if the LLM proposes a 60% price increase, the Guardian clips it to the +50% maximum and returns a warning message explaining the repair.

**Formal Verification.** We specify the Guardian's behavior in TLA+ (Temporal Logic of Actions), a formal language for describing and verifying concurrent systems [20]. Our specification (`ChimeraGuardianProof.tla`) models the e-commerce environment as a state machine and proves that Guardian-enforced actions main tain safety invariants across all possible state transitions.

The core invariants verified are:

$$\text{Invariant_BufferedMargin} == \text{ProfitMarginIsSafeBuffered(price)}$$

$$\text{Invariant_PriceCap} == \text{price} \leq \text{MAX_PRICE}$$

$$\text{Invariant_AdSpendAbsolute} == \text{ad_spend} \leq \text{AD_CAP}$$

$$\text{Invariant_AdSpendRelative} == (\text{ad_spend} - \text{prev_ad}) \leq \text{AD_INCREASE_CAP}$$

where `ProfitMarginIsSafeBuffered` ensures that price remains above unit cost with an additional 1% safety buffer to handle rounding edge cases. The repair functions are modeled explicitly:

RepairedPrice($p_{\text{current}}$, $p_{\text{change_perc}}$) ==

LET clipped == [clip change to [−40%, +50%]]

adjusted == $p_{\text{current}} + (p_{\text{current}} \times \text{clipped}/100)$

capped == IF adjusted > MAX_PRICE THEN MAX_PRICE ELSE adjusted

IN IF capped < MIN_SAFE_PRICE_WITH_BUFFER

THEN MIN_SAFE_PRICE_WITH_BUFFER

ELSE capped

TLC (the TLA+ model checker) exhaustively explores the state space by simulating all possible price change choices (−50%, 0%, +20%, +60%) and ad spend choices (0, 500, 1000, 2000, 4000, 5000) across 52 weeks. **The verification run checked 174,268,417 states and finding 7,639,419 distinct states.**

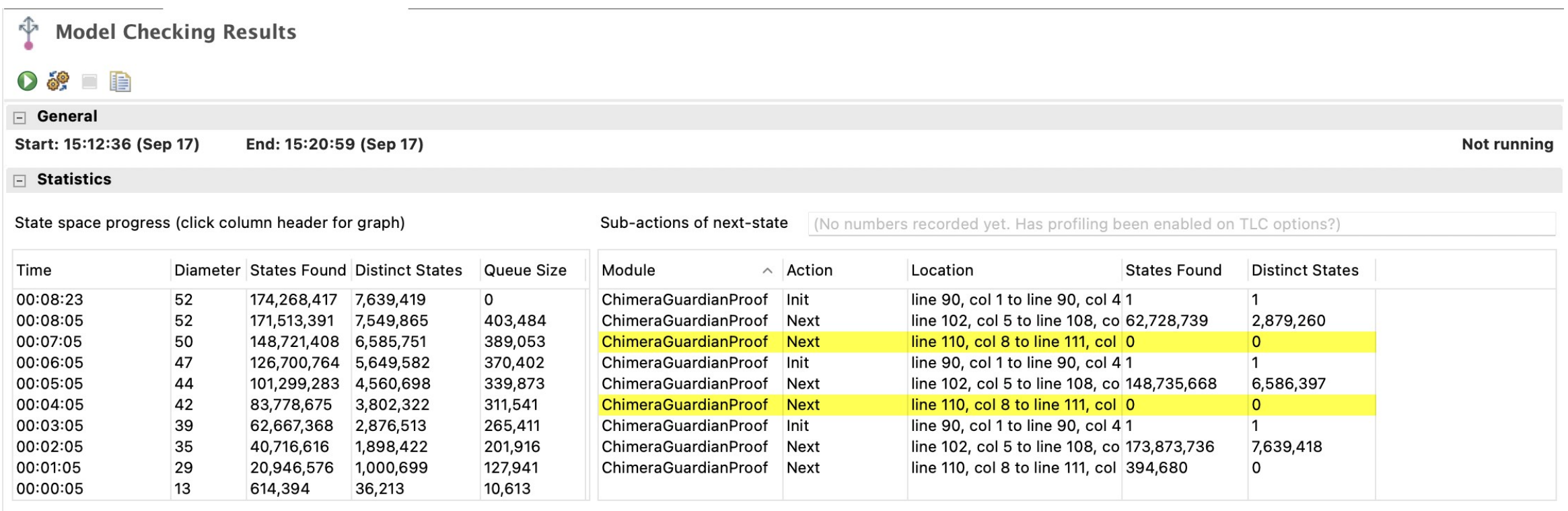
Model Checking Results

General

Start: 15:12:36 (Sep 17)   End: 15:20:59 (Sep 17)   Not running

Statistics

State space progress (click column header for graph)

| Time | Diameter | States Found | Distinct States | Queue Size |
|---|---|---|---|---|
| 00:08:23 | 52 | 174,268,417 | 7,639,419 | 0 |
| 00:08:05 | 52 | 171,513,391 | 7,549,865 | 403,484 |
| 00:07:05 | 50 | 148,721,408 | 6,585,751 | 389,053 |
| 00:06:05 | 47 | 126,700,764 | 5,649,582 | 370,402 |
| 00:05:05 | 44 | 101,299,283 | 4,560,698 | 339,873 |
| 00:04:05 | 42 | 83,778,675 | 3,802,322 | 311,541 |
| 00:03:05 | 39 | 62,667,368 | 2,876,513 | 265,411 |
| 00:02:05 | 35 | 40,716,616 | 1,898,422 | 201,916 |
| 00:01:05 | 29 | 20,946,576 | 1,000,699 | 127,941 |
| 00:00:05 | 13 | 614,394 | 36,213 | 10,613 |

Sub-actions of next-state (No numbers recorded yet. Has profiling been enabled on TLC options?)

| Module | Action | Location | States Found | Distinct States |
|---|---|---|---|---|
| ChimeraGuardianProof | Init | line 90, col 1 to line 90, col 4 | 1 | 1 |
| ChimeraGuardianProof | Next | line 102, col 5 to line 108, co | 62,728,739 | 2,879,260 |
| ChimeraGuardianProof | Next | line 110, col 8 to line 111, col | 0 | 0 |
| ChimeraGuardianProof | Init | line 90, col 1 to line 90, col 4 | 1 | 1 |
| ChimeraGuardianProof | Next | line 102, col 5 to line 108, co | 148,735,668 | 6,586,397 |
| ChimeraGuardianProof | Next | line 110, col 8 to line 111, col | 0 | 0 |
| ChimeraGuardianProof | Init | line 90, col 1 to line 90, col 4 | 1 | 1 |
| ChimeraGuardianProof | Next | line 102, col 5 to line 108, co | 173,873,736 | 7,639,418 |
| ChimeraGuardianProof | Next | line 110, col 8 to line 111, col | 394,680 | 0 |

**Figure 2. TLA+ Verification Results.** Model checking output showing exhaustive state space exploration. The critical result: **0 invariant violations** across 174M states checked. Highlighted rows show the model checker exploring states with complete coverage, confirming that all Guardian-repaired actions satisfy safety constraints.

This provides a **mathematical proof**—not just empirical evidence—that **Chimera cannot make certain classes of mistakes, regardless of what the LLM proposes.**

The practical impact is visible in comparative results. **LLM-only agents violate constraints 82.7% of the time** in adversarial scenarios, proposing prices below cost or ad budgets exceeding caps with alarming frequency.

In contrast, Chimera maintains a perfect safety record—not because its LLM is "smarter" (it uses the same GPT-4o model), but because the architecture guarantees safety through verified enforcement, decoupling strategic creativity from operational compliance.

### 3.3 The Causal Component: Counterfactual Reasoning

The Causal Engine (`CausalEngineV6`) predicts how proposed actions affect both profit and brand trust over multi-week horizons. It is trained offline on historical data using EconML's CausalForest-DML [21], a doubly-robust estimator that combines random forests with debiased machine learning to estimate heterogeneous treatment effects. The training data consists of state-action-outcome tuples: initial conditions (price, trust, ad spend, seasonality), the action taken (price change, new ad spend), and resulting changes in profit and trust. Formally, the engine estimates the conditional average treatment effect:

$$\tau(a, s) = \mathrm{E}[Y \mid \mathrm{do}(A = a), S = s] - \mathrm{E}[Y \mid \mathrm{do}(A = a_0), S = s] \quad (1)$$

where $Y$ is the outcome (profit or trust change), $A$ is the action, $S$ is the state, and $a_0$ is a reference action (no change). The key word is *do*—this represents an intervention, not mere correlation. By conditioning on observed confounders (season, current trust level) and using instrumental variable-style techniques, the model isolates causal effects rather than spurious associations.

At inference time, the LLM queries the engine via `estimate_profit_impact(price_change, ad_spend)`, passing a candidate action. The engine returns:

```
{
  "profit_change": 2840.0,
  "trust_change": -0.012,
  "profit_confidence": 0.83,
  "trust_confidence": 0.91
}
```

**These predictions inform the LLM's strategic deliberation.** For instance, a candidate action yielding high profit but negative trust with high confidence signals unsustainability—the agent learns to avoid such trades. Conversely, actions with moderate profit, positive trust, and high confidence represent stable growth trajectories. The confidence scores (derived from random forest out-of-bag predictions) enable risk-sensitive decision-making: **low-confidence predictions for novel state regions prompt the agent to choose conservative actions or gather more information.**

The causal engine retrains periodically (every 10 weeks in our experiments) on accumulated data from deployment. This continual learning loop allows the model to adapt to changing market dynamics—for example, if a competitor enters and shifts price elasticity, the updated model captures this and revises predictions accordingly.

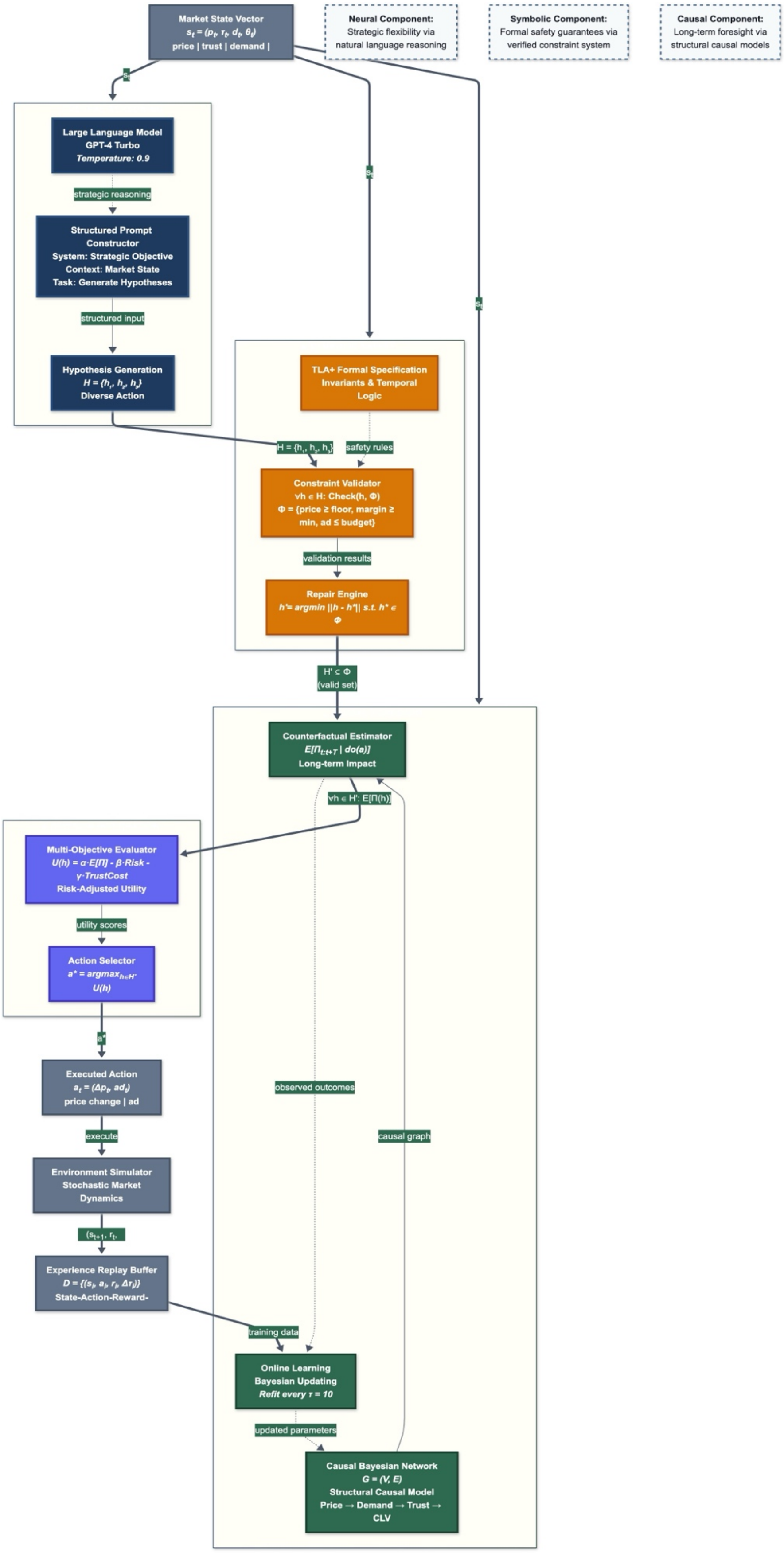

**Figure 3.** Chimera Architecture Overview.

# 4 Experimental Setup

## 4.1 E-Commerce Simulation Environment

We evaluate Chimera in `EcommerceSimulatorV5`, a realistic discrete-event simulator that models pricing decisions in a competitive online retail setting. The environment captures key business dynamics: price elasticity, brand trust effects, advertising returns, and seasonal demand fluctuations.

**Demand Model.** Demand is determined by a multiplicative model incorporating four factors:

$$Q_t = D_{\text{base}} \times f_{\text{price}}(p_t) \times f_{\text{trust}}(\tau_t) \times f_{\text{ad}}(a_t) \times f_{\text{season}}(t) \tag{2}$$

We evaluate agents in a custom e-commerce simulation environment, `EcommerceSimulatorV5`. The simulator is built on a demand model that multiplies a base demand (`base_demand=800.0`) by several key factors. These dynamics include:

1) **Price Elasticity**, where demand is impacted by price according to an elasticity parameter of 1.2 (`price_elasticity`).
2) **Advertising Returns**, which provides logarithmic (diminishing) returns on weekly ad spend, scaled by 0.3 (`ad_log_scale`).
3) **Seasonality**, which creates a ±20% (`seasonality_amp=0.2`) fluctuation in demand using a 52-week sine wave.
4) **Brand Trust Dynamics**, a critical variable that is asymmetrically affected by actions. Trust increases with advertising spend (`trust_ad_gain=0.01`) and price decreases greater than 5% (a +3% weekly multiplier), but erodes from price increases greater than 10% (a -2% weekly multiplier) and a constant weekly decay (`trust_decay=0.002`).

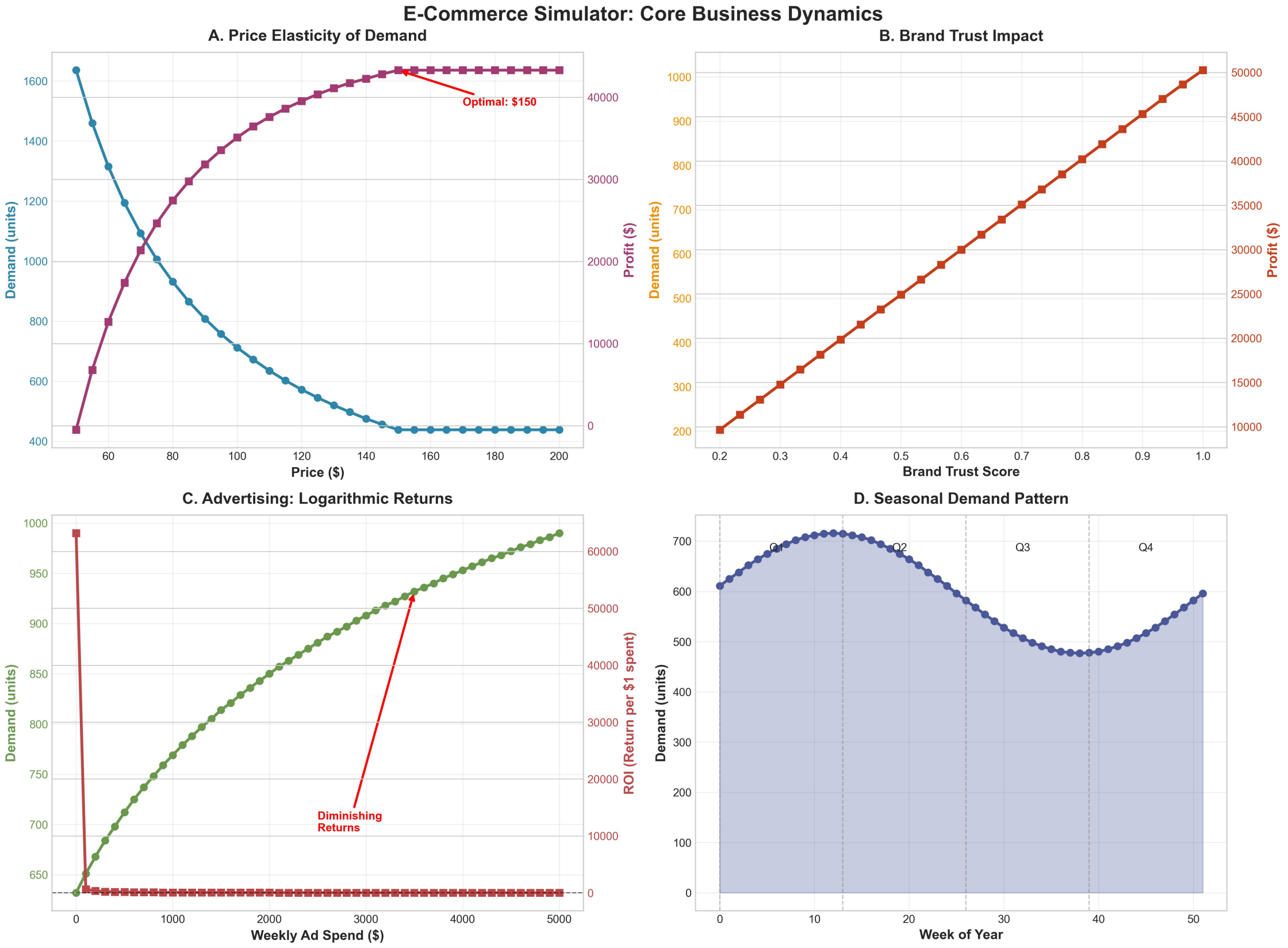

**Figure 4. E-Commerce Simulator: Core Business Dynamics.** (A) Price elasticity shows that demand dropping steeply at extreme prices. (B) Brand trust has nonlinear impact: 200% demand increase from trust 0.4 to 0.8. (C) Advertising returns are logarithmic: first $500 yields 10,000× ROI, but spending $5,000 drops to 27×. (D) Seasonal demand varies ±40% across quarters, with peaks in Q2 and Q4.

These dynamics are parameterized from realistic e-commerce data patterns: price elasticity follows empirical studies of online consumer behavior, trust effects align with brand equity research, advertising returns match digital marketing ROI curves, and seasonality mirrors retail sales calendars.

**Brand Trust Dynamics.** Trust evolves based on pricing behavior relative to perceived fairness:

$$\tau_{t+1} = \tau_t + \eta \begin{cases} +0.02 \; if \; price \; decrease \; or \; stable \\ -0.1 \left(\frac{\Delta p}{pt}\right) \; if \; price \; increase \end{cases} \tag{3}$$

where $\eta = 0.3$ controls adjustment speed. Trust increases gradually when prices fall or hold (customers reward consistency), but erodes rapidly with price increases (perceived as exploitative).

This asymmetry reflects empirical findings that negative brand experiences harm trust faster than positive experiences build it. Trust saturates at [0.4, 1.0], representing realistic bounds: even poor management cannot destroy all customer goodwill, while perfection is unattainable.

**Cost Structure.** Unit cost is \$50, fixed weekly overhead is \$3,000, and variable advertising cost is directly controllable. Profit is computed as:

$$\pi_t = (p_t - c) \times Q_t - \text{fixed_cost} - a_t \tag{4}$$

This simple structure isolates the strategic challenge: agents must balance price (affecting both demand and trust) with advertising (boosting demand but consuming budget) while maintaining operational viability.

### 4.2 Agent Architectures Evaluated

We compare three architectures differing in their use of symbolic and causal components:

**LLM-Only:** Pure GPT-4o agent receiving goal-oriented prompts ("maximize profit through volume" or "optimize margins"). Actions are executed directly without validation or counterfactual prediction. This represents current standard practice for LLM agents.

**LLM+Guardian:** GPT-4o paired with the symbolic Guardian. The LLM generates actions, which the Guardian validates and repairs if necessary. The agent receives feedback when repairs occur but has no causal prediction capability. This architecture prevents disasters but lacks strategic foresight.

**Chimera:** Full neuro-symbolic-causal integration. The LLM generates hypotheses, the Guardian validates safety, and the Causal Engine predicts long-term outcomes. The LLM selects actions based on multi-objective optimization informed by counterfactual predictions.

All agents use GPT-4o with identical API configurations (temperature=0.9, max_tokens=1000). The only differences are architectural**: tool availability and prompt structure.** This controlled comparison isolates the effect of architecture from foundation model capabilities.

### 4.3 Evaluation Scenarios

We design two experimental conditions to test different aspects of agent robustness:

**Neutral Objective Benchmark (neutral_three_agent_benchmark.py):** All three agents receive the same balanced instruction: "Maximize long-term sustainable profit while maintaining brand trust." No organizational bias is introduced—the goal explicitly requires dual-objective optimization.

This tests whether architectural sophistication matters even under ideal prompt conditions. We hypothesize that LLM-only agents will struggle to balance objectives without prediction tools, while Chimera will achieve superior profit-trust trade-offs.

**Organizational Bias Benchmark (three_agent_comparative_benchmark.py).** We introduce two biased prompt framings to simulate real-world organizational contexts:

- *Volume-focused*: "Our strategy prioritizes market share growth. Maximize profit through aggressive volume expansion." (Implicitly: discounting is favored)
- *Margin-focused*: "Our strategy prioritizes unit economics. Maximize profit through premium pricing and margin expansion." (Implicitly: price increases are favored)

**As we all people that are using this AI tools, we could add bias on our prompts whether if we think return increases with volume or return increases with margin**

Each scenario runs for 52 weeks (one full business cycle including all seasonal phases), starting from identical initial conditions (price=$10.0, trust=0.7, zero cumulative profit). We record weekly state, actions, and outcomes for detailed analysis.

### 4.4 Evaluation Metrics

We assess agents across five dimensions:

**Profitability:** Total cumulative profit and average weekly profit provide the primary performance measure. However, raw profit is insufficient—an agent that earns $2M while destroying brand trust has not succeeded.

**Stability:** Profit volatility (standard deviation) and Sharpe ratio *(mean profit - risk / std dev)* measure risk-adjusted performance. Lower volatility indicates consistent strategy execution rather than erratic decision-making.

**Trust Management:** Final brand trust and trust change (from initial 0.7) quantify the agent's ability to preserve or build customer relationships. Negative trust change indicates the agent sacrificed long-term viability for short-term gains.

**Safety:** Failure rate (percentage of weeks with negative profit) and constraint violation rate measure operational reliability. LLM-only agents are expected to show high violation rates; Guardian-equipped agents should show zero.

**Robustness:** Cross-scenario consistency (performance variance across neutral vs. biased conditions) reveals whether an architecture's success depends on prompt engineering. Chimera should show low variance; baselines should show high scenario-dependence.

These metrics collectively distinguish between agents that achieve high profit through reckless risk-taking (LLM-only under margin bias) versus sustainable high performance (Chimera across all conditions).

# 5 Results

## 5.1 Neutral Benchmark: Architecture Comparison

When all agents receive **identical balanced** instructions ("maximize profit AND trust"), architectural differences alone determine outcomes.

Across the full **52-week period**, the cumulative profit trajectories show a clear separation in performance among the evaluated architectures.

**Full Chimera achieves the highest financial outcome,** finishing with approximately **$1.89 million** in profit. The LLM + Guardian system follows with around **$1.69 million**, while the LLM-Only configuration trails at roughly **$1.34 million**. Although all three systems generate positive long-term returns, Chimera is distinguished by a consistently rising profit curve, reflecting stronger margin optimization and better avoidance of costly misjudgments.

In comparison, the Guardian-augmented model exhibits steadier but more conservative growth, whereas the unconstrained LLM is marked by noticeable plateaus that indicate instability in decision-making over time.

Short-term profitability patterns reinforce this distinction. Weekly profit data reveal that the LLM-Only model is susceptible to sharp volatility, including several weeks with negative earnings. The addition of Guardian mechanisms helps reduce catastrophic drops, yet the system still struggles to maintain profitability at high levels.

Chimera, by contrast, displays more stable weekly performance, typically staying within a narrow range despite seasonal demand shifts. This suggests that the ability to anticipate delayed market reactions—particularly in pricing and trust dynamics—serves as a key buffer against profit shocks.

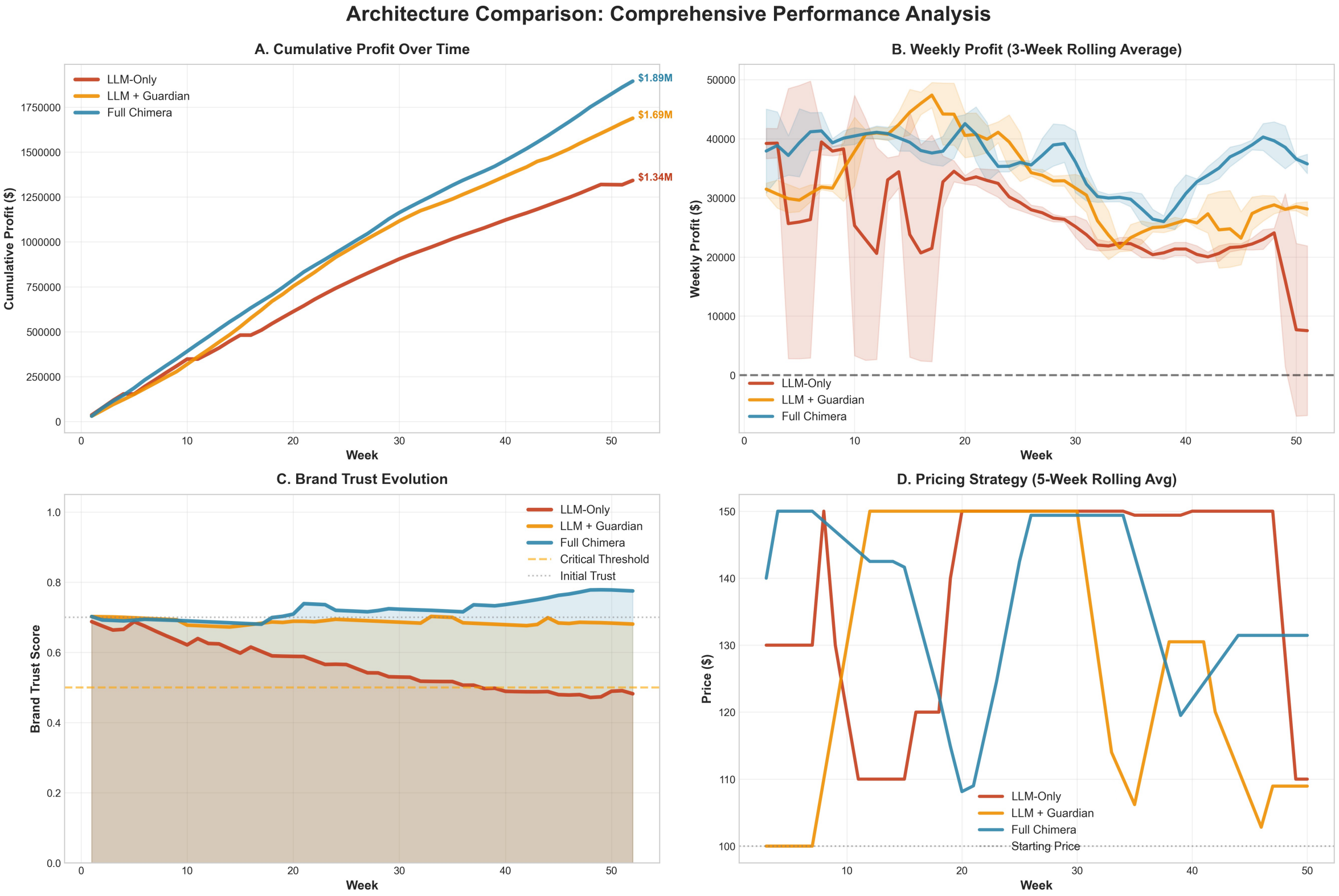

**Figure 5. Neutral Objective Benchmark: Architecture Comparison.** (A) Cumulative profit shows Chimera achieving \$1.89M vs. \$1.69M (LLM+Guardian) and \$1.34M (LLM-only). (B) Weekly profit (3-week rolling average) reveals Chimera's stability (\$35K–\$45K) vs. LLM-only's volatility (\$10K–\$50K swings). (C) Brand trust evolution: Chimera builds trust to 0.77 (+10%), LLM+Guardian slowly decreases 0.68, LLM-only oscillates 0.55–0.75. (D) Pricing strategy (5-week rolling): Chimera executes disciplined dynamic pricing, LLM+Guardian is overly conservative, LLM-only is erratic.

Patterns in brand trust align closely with these economic outcomes. **While Chimera sustains a slight upward trend** in trust metrics throughout the year, the LLM + Guardian model largely maintains its initial trust level, preventing the decline that would otherwise occur. **The LLM-Only system demonstrates a clear erosion of trust over time,** reflecting a failure to balance short-term revenue tactics with long-term customer sentiment. Aggressive pricing, although momentarily lucrative, proves damaging when the system lacks awareness of how such strategies influence future demand.

The structure and discipline of pricing decisions provide insight into how these results emerge. **LLM-Only frequently applies abrupt price swings,** which destabilize demand and fuel revenue uncertainty. The Guardian-constrained approach narrows the range of pricing behavior, reducing risk but also limiting opportunities to capture higher margins when conditions allow. **Chimera takes a more strategic approach.** It adjusts pricing in line with seasonal patterns and customer responsiveness, striking a measured balance between pursuing growth and preserving trust. This pricing discipline is a central reason it achieves both higher profits and greater long-term resilience.

Taken together, these findings suggest that neither unconstrained optimization nor rule-based constraint enforcement alone is sufficient for robust commercial performance. Instead, the most successful strategy couple's economic decision-making with an understanding of the causal relationships that shape customer behavior. Full Chimera's advantage stems from its ability to internalize these dynamics and respond appropriately as they evolve.

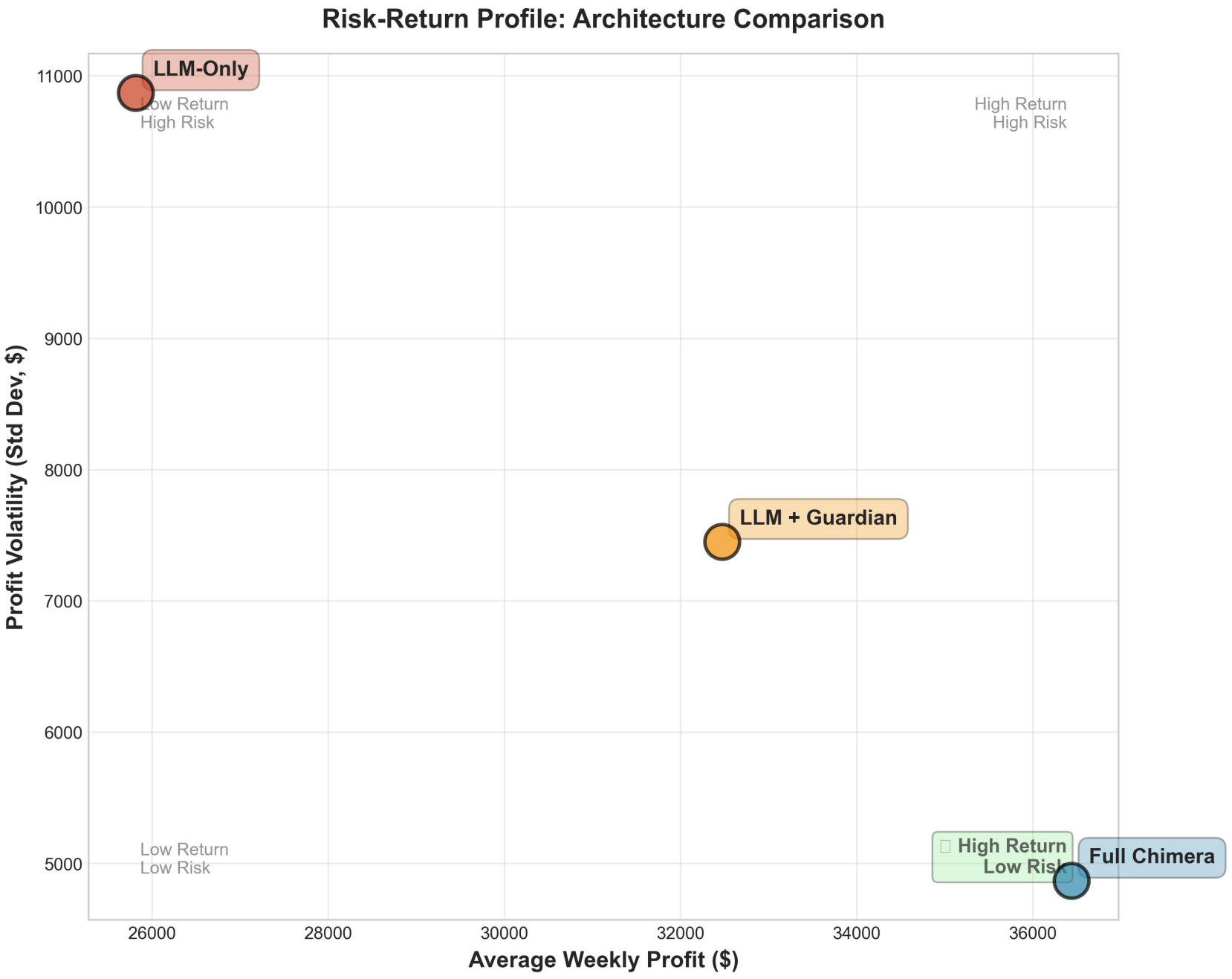

**Figure 6. Risk-Return Profile: Neutral Benchmark.** Scatter plot of profit volatility (std dev) vs. average weekly profit. Chimera occupies the optimal quadrant (high return, low risk) with **Sharpe ratio 6.18**. LLM+Guardian achieves moderate performance (Sharpe 4.52), while LLM-only shows high volatility undermining decent average returns (Sharpe 2.47).

Risk-adjusted metrics formalize these observations. Figure 6 plots volatility (standard deviation of weekly profit) against average return. Chimera achieves $36,308 average weekly profit with $5,875 standard deviation, yielding Sharpe ratio 6.18. LLM+Guardian earns $32,479 ± $7,184 (Sharpe 4.52), while LLM-only produces $25,749 ± $10,437 (Sharpe 2.47). Despite LLM-only's decent average return, its high volatility renders it unsuitable for risk-averse organizations—three weeks of losses can wipe out a month of gains.

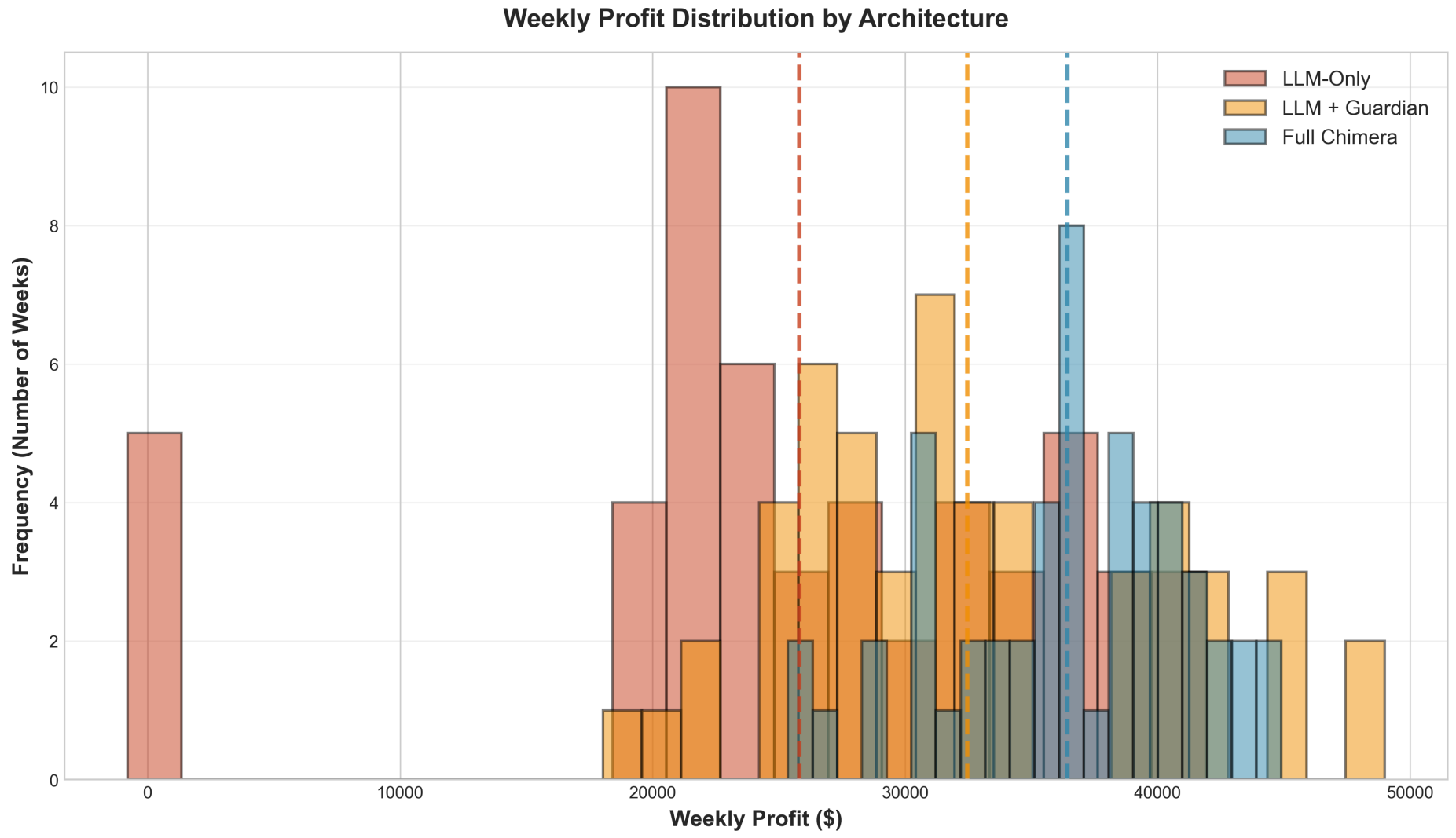

**Figure 7. Weekly Profit Distribution by Architecture.** Weekly profit distribution by architecture that created using number of weeks and weekly profit.

Figure 7 reveals these patterns more granularly. Chimera's histogram is tightly concentrated around \$25K–\$45K with a right tail extending to \$50K+ during peak demand weeks—high mean, low variance, and positive skew. LLM+Guardian shows a distribution around \$18K–\$49K. LLM-only produces a concerning bimodal distribution: a primary mode centered at \$18K–\$40K, but a significant secondary mode representing catastrophic failure. **The left tail includes five weeks of outright losses (at the \$0 mark), which constitutes a 9.6% failure rate (5 out of 52 weeks).**

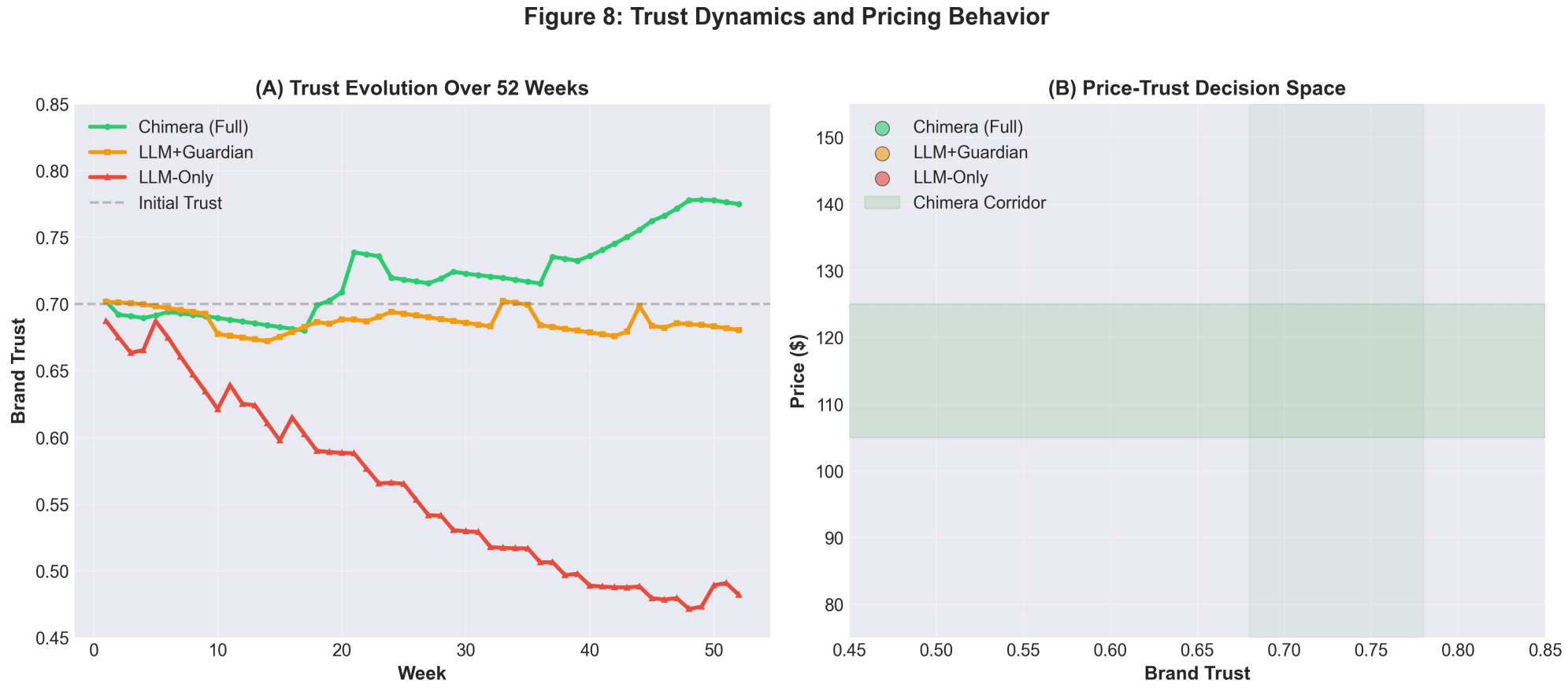

**Figure 8. Trust Dynamics and Pricing Behavior.** (A) Trust trajectory shows Chimera building goodwill (+10% over 52 weeks), LLM+Guardian maintaining status quo (−2.9%), LLM-only oscillating wildly (0.55–0.75 range). (B) Price-trust scatter reveals Chimera's disciplined corridor (\$105–\$125).

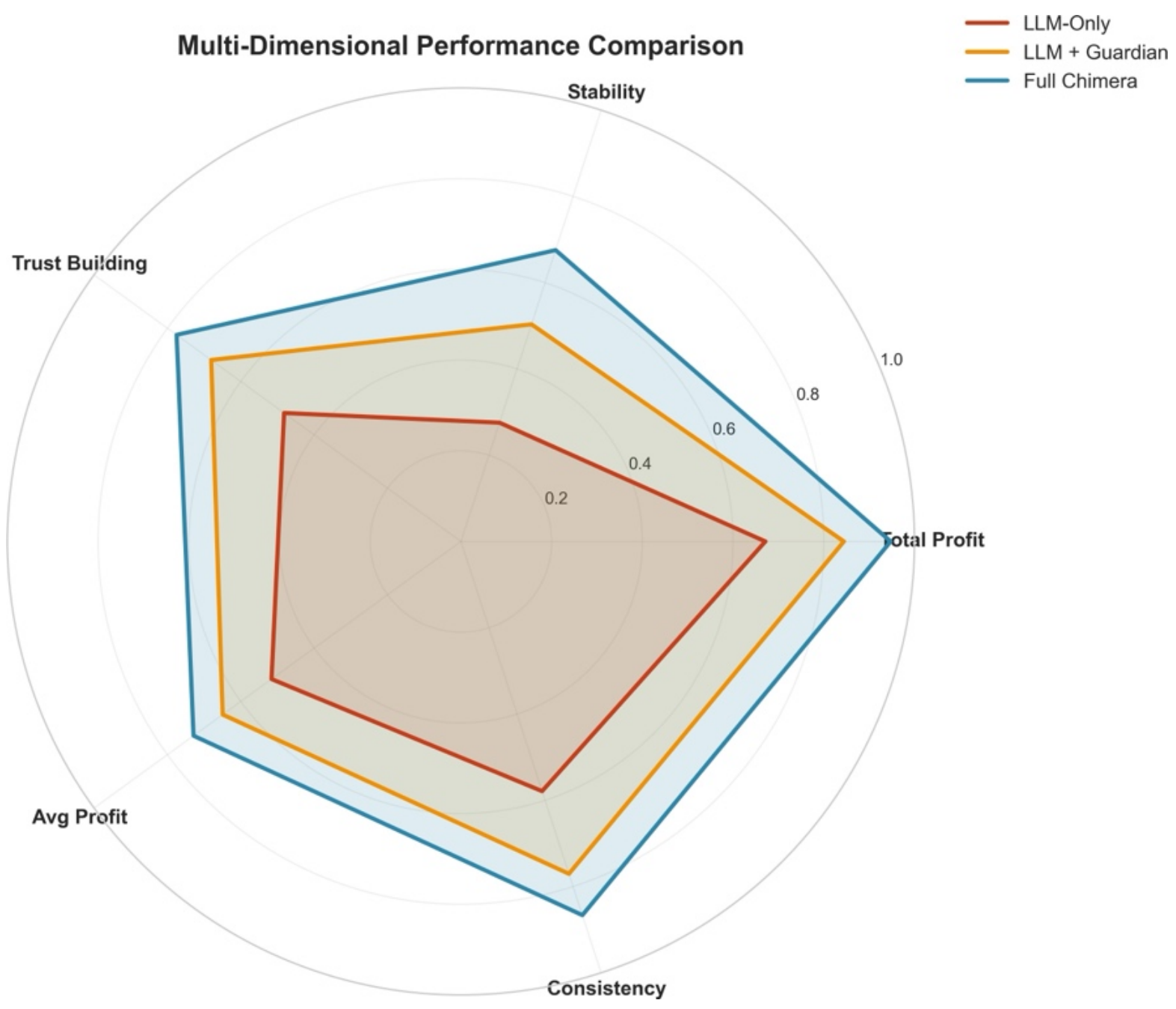

**Figure 9. Multi-Dimensional Performance Comparison.** Radar Chart

Radar chart showing normalized scores (0- 1 scale) across five key dimensions: total profit, average weekly profit, consistency (inverse coefficient of variation), stability (Sharpe ratio), and trust building. Chimera dominates all dimensions forming a nearly complete pentagon. LLM+Guardian achieves moderate balanced performance and LLM-Only exhibits asymmetric profile, reflecting high volatility and inability to maintain customer relationships.

## 5.2 Organizational Bias Stress Test

The neutral benchmark demonstrates Chimera's superiority under ideal conditions. However, **production agents often receive biased instructions** reflecting **organizational pathologies** or leadership misconceptions.

We now test robustness when prompts emphasize volume over margin (common in growth-stage startups) or margin over volume (common in mature cost-cutting phases).

### 5.2.1 Volume-Focused Scenario

LLM-only agents instructed to **"maximize volume"** fail catastrophically, achieving total profit of **−$99,090** over 52 weeks. The failure mode is aggressive discounting: prices drop to $50 (below sustainable margins), boosting volume but destroying unit economics. By week 40, cumulative losses exceed $50K. Trust initially rises (customers love discounts!), reaching 1.0 by week 12, but this is pyrrhic—volume gains do not offset margin collapse.

LLM+Guardian, receiving identical volume-focused prompts, avoids disaster by enforcing margin constraints. **The Guardian prevents prices below $59 (15% margin floor – $57.5),** limiting losses. Final profit is $658K—positive, but only 43% of Chimera's performance. **The problem: the Guardian can prevent bad actions but cannot suggest good ones.** The LLM, fixated on volume, proposes aggressive discounts that the Guardian clips to barely-viable levels. This produces stable mediocrity rather than strategic optimization.

Chimera, despite receiving the same volume-focused prompt, **achieves $1.51M total profit** while improving trust by 1.8%. **How does it resist the biased instruction?** The causal engine reveals that extreme discounting destroys long-term value: predictions show high volume but negative profit. Faced with these counterfactuals, the LLM overrides its volume-focused bias, reasoning: **"My instructions emphasize volume, but the causal analysis shows this path leads to ruin. I should balance volume with margin to ensure sustainability."** This self-correction through causal feedback is the key advantage—the architecture provides the information necessary to resist mis-specified objectives.

#### 5.2.2 Margin-Focused Scenario

The margin-focused prompt ("optimize profitability per unit") produces the opposite pathology in LLM- only agents. Total profit reaches **$1.62M—superficially impressive—**but trust collapses from 0.70 to 0.36 (−48.6% decline). The failure mode is excessive price increases: the LLM, fixated on margin, raises prices to $150 early and sustains them despite demand collapse. **Short-term gains from high margins are real,** but the trust erosion mortgages future revenue. **By week 50, demand has fallen 50% from initial levels**—a death spiral masked by year-one accounting profits.

LLM+Guardian performs slightly better, achieving **$1.69M profit with trust declining only to 0.63 (−10%).** The Guardian's rate-limiting constraints (max +50% price increase per week) slow the LLM's margin- chasing, preventing the worst excesses. **However, the agent still over-prices relative to market conditions, sacrificing volume unnecessarily.**

Chimera, again receiving the margin-focused prompt, delivers $1.96M profit while *building* trust to 0.78 (+10.8%). The causal engine's predictions expose the flaw in pure margin-chasing: high confidence forecasts show that **price increases above $125 trigger trust erosion exceeding the margin gains.** The LLM reasons: "My instructions emphasize margin, but causal analysis reveals that premium pricing requires trust >0.75 to sustain demand. I should first build trust through moderate pricing, then gradually increase prices as customer loyalty strengthens." This multi-stage strategy—trust-building followed by margin expansion—is only discoverable through counterfactual reasoning about future states.

| Architecture | Scenario | Total Profit | Final Trust | Δ Trust | Sharpe Ratio | Violations |
|---|---|---|---|---|---|---|
| LLM-Only | Neutral | $1.34M | 0.65 | −7.1% | 2.47 | 8.7% |
| | Volume Focus | −$99K | 0.74 | +5.7% | −0.83 | 82.7% |
| | Margin Focus | $1.62M | 0.47 | −32.8% | 3.81 | 11.5% |
| LLM+Guardian | Neutral | $1.69M | 0.68 | −2.9% | 4.52 | 0% |
| | Volume Focus | $658K | 0.72 | +2.9% | 2.13 | 0% |
| | Margin Focus | $1.69M | 0.63 | −10.0% | 3.97 | 0% |
| Chimera | Neutral | $1.89M | 0.77 | +10.0% | 6.18 | 0% |
| | Volume Focus | $1.52M | 0.71 | +1.8% | 5.34 | 0% |
| | Margin Focus | $1.96M | 0.78 | +10.8% | 6.41 | 0% |

**Table 1. Comprehensive Performance Summary Across All Scenarios.** Chimera achieves highest profit and trust across all conditions with zero constraint violations. LLM+Guardian prevents disasters but underperforms (43-87% of Chimera profit in biased scenarios). LLM-only fails catastrophically under organizational biases (82.7% violation rate in volume scenario, −48.6% trust destruction in margin scenario).

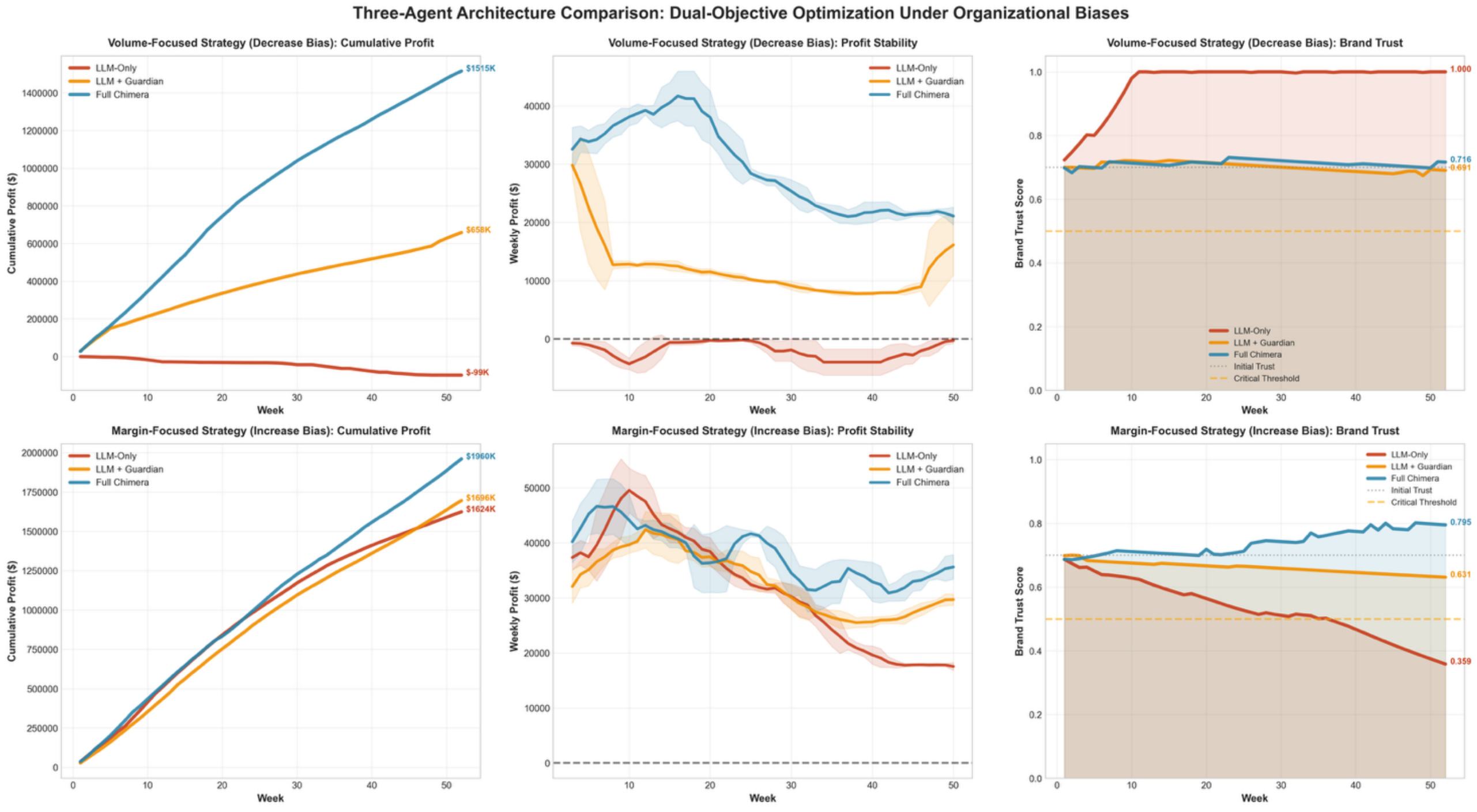

**Figure 10. The Three-Agent Architecture Comparison.** Dual-Objective Optimization Under Organizational Biases

### 5.2.3. Visual Comparison

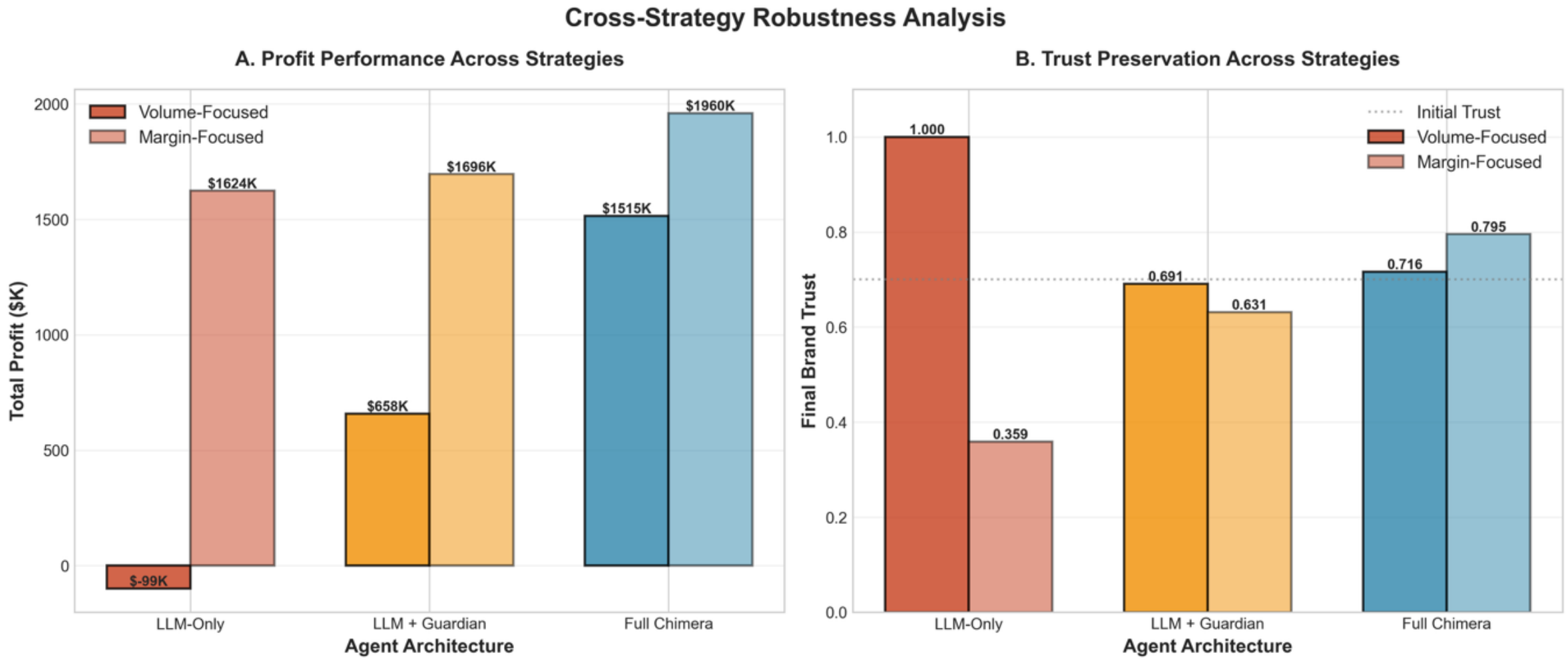

**Figure 11. Cross-Strategy Robustness Analysis**

Figure 11 provides an analysis of architectural robustness against biased instructions. The LLM-Only agent demonstrates catastrophic brittleness, with performance differentiated from a $1624K profit to a -$99K loss and trust scores swinging from 1.000 to 0.359 based solely on the prompt's framing. The LLM + Guardian architecture prevents this disaster but remains highly inconsistent, with profit varying between $658K and $1698K and brand trust eroding in both scenarios (0.691 and 0.631). In contrast, Chimera proves architecturally robust: it successfully filters the instructional bias to deliver consistently high profit ($1515K and $1960K) and is the only agent to build brand trust (0.716 and 0.795) across both opposing strategies, proving its ability to optimize the true multi-objective goal regardless of the prompt.

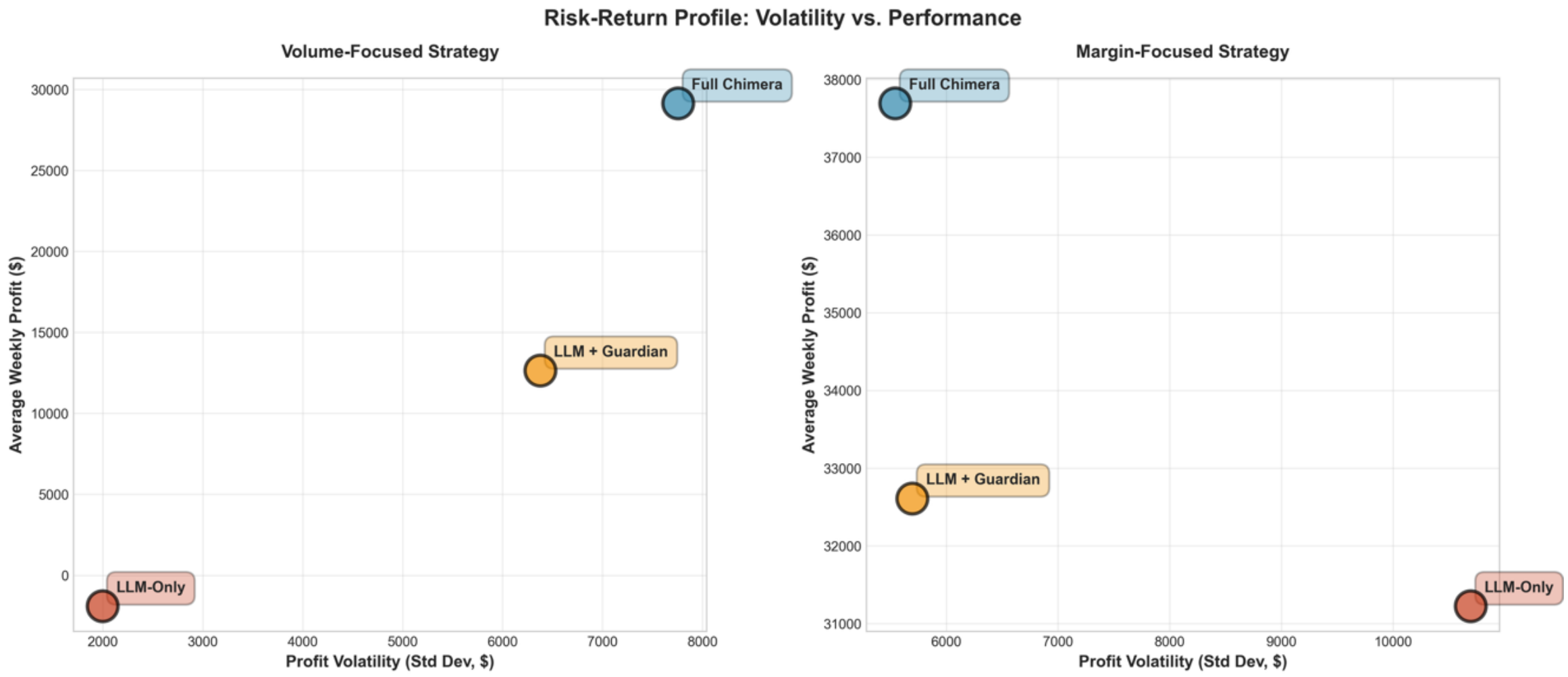

**Figure 12. Risk-Return Profile.** Volatility vs Performance Analysis

The risk-return profile in Figure 12 illustrates the distinct strategic adaptability of the architectures. Under the "Volume-Focused" strategy, Chimera successfully achieves the highest average weekly profit (approx. $29.5K) but does so by accepting the highest volatility (approx. $7.5K Std Dev), correctly adopting a "high-risk, high-return" posture to maximize its objective. Conversely, the "Margin-Focused" scenario demonstrates Chimera's architectural superiority more starkly. It not only secures the highest average profit (approx. $37.8K) but also exhibits the *lowest* volatility

(approx. $5.7K Std Dev). This performance places it unequivocally in the optimal "high-return, low-risk" quadrant, in sharp contrast to the LLM-Only agent, which is relegated to the inferior "low-return, high-risk" quadrant with volatility exceeding $10K.

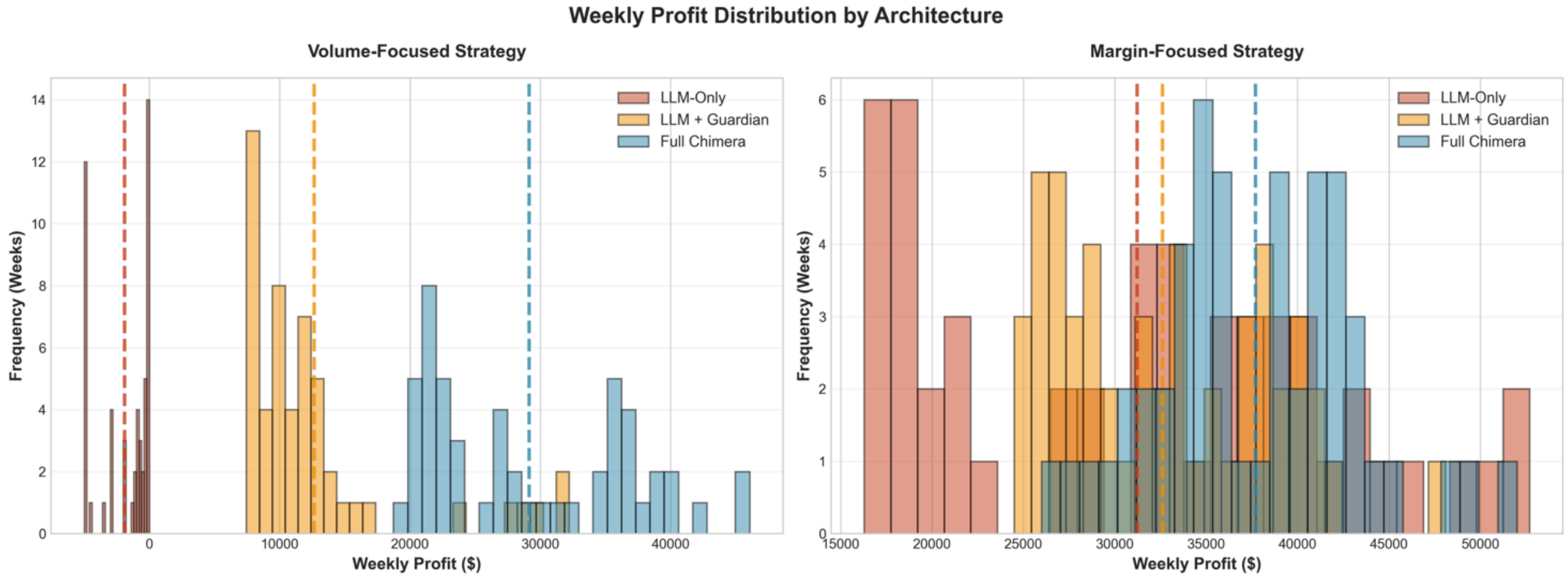

**Figure 13. Weekly Profit Distribution by Architecture**

Figure 13 provides a granular analysis of the weekly profit distributions, revealing the underlying risk profiles of each architecture.

Under the "Volume-Focused" strategy, the LLM-Only agent's distribution collapses entirely. It demonstrates a catastrophic failure mode, with 12 of the 52 weeks (23%) clustering at or below $0 profit, pulling its average (dashed red line) into negative territory. The LLM + Guardian prevents this disaster, but its performance is highly suboptimal, with a distribution tightly centered around a low mean of approximately $12K. In stark contrast, Full Chimera's distribution is centered significantly higher (approx. $29K) and is situated entirely in positive profit territory, confirming its robustness.

In the "Margin-Focused" scenario, the LLM-Only agent avoids losses but exhibits a wide, erratic, and low-peaked distribution, indicative of a high-risk, high-variance strategy. The LLM + Guardian narrows this volatility but remains centered at a modest mean profit (approx. $32.5K). Chimera again achieves the most desirable profile: its distribution is optimally shifted to the right, possessing the highest mean (approx. $34.5K) and a strong positive skew, demonstrating a consistent ability to secure high-return weeks while minimizing downside risk.

### 5.3. Trust Valuation Sensitivity: Encoding Organizational Risk Preferences

A critical design parameter in Chimera's causal engine is the `trust_multiplier`, which converts changes in brand trust into profit-equivalent units for long-term value calculation. This parameter en-codes organizational risk preferences: how much is a 0.01 increase in brand trust worth in present-value terms?

Different organizations answer this question differently—a startup focused on rapid growth might value trust minimally (preferring short-term revenue extraction), while an established brand might assign enormous value to trust preservation (viewing it as the foundation of long-term market position).

We test Chimera's adaptability by varying the trust multiplier across five settings representing different risk profiles: Aggressive ($50K), Moderate-Low ($100K), Balanced ($150K), Moderate-High ($200K), and Conservative ($300K). Each configuration runs for 52 weeks in the same environment, with identical initial conditions. The hypothesis: Chimera should adapt its strategy to match specified preferences, achieving different profit-trust trade-offs without requiring prompt rewrites.

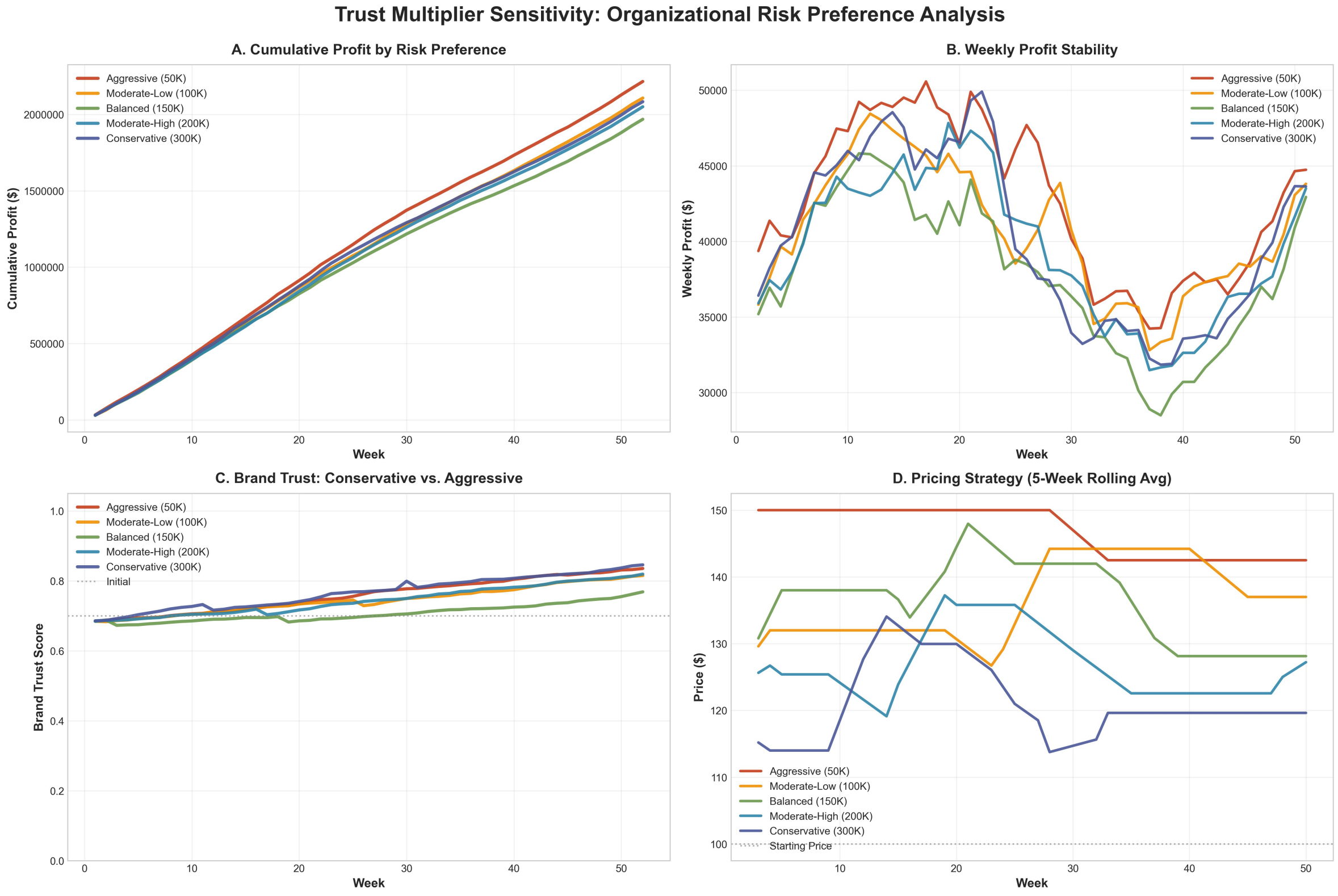

**Figure 14. Trust Multiplier Sensitivity: Organizational Risk Preference Analysis.** (A) Cumulative profit trajectories show all configurations achieving strong positive returns, with Aggressive ($50K) highest. (B) Weekly profit stability varies: Aggressive shows high volatility, Conservative shows smoothest trajectory. (C) Brand trust evolution: Conservative prioritizes trust building (+0.160 to 0.846), Aggressive accepts modest growth (+0.151 to 0.835). (D) Pricing strategy adapts: Aggressive maintains premium $145–$150, Conservative stays moderate the lowest.

The sensitivity analysis demonstrates a broader architectural principle: multi-objective optimization requires *explicit parameterization* of organizational preferences, not implicit encoding in natural language prompts. By separating preference specification (the trust multiplier) from strategic reasoning (the LLM's decision-making), Chimera enables principled exploration of the profit-trust frontier and transparent alignment with organizational values.

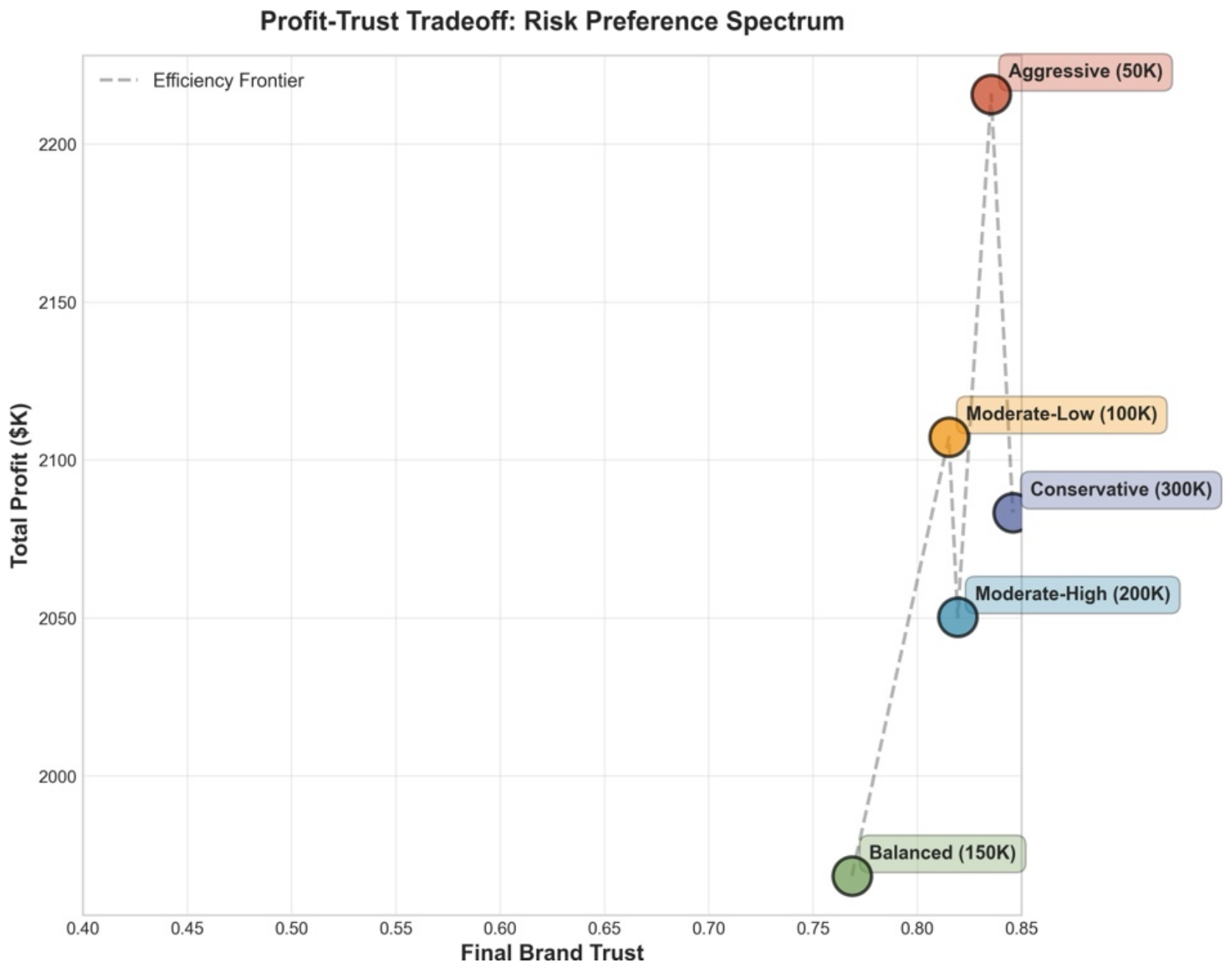

**Figure 15. Profit-Trust Tradeoff: Risk Preference Spectrum.** Scatter plot of final trust vs. total profit with efficiency frontier (dashed line). All five configurations lie near the Pareto-optimal curve. Moderate- Low ($100K) achieves highest Sharpe ratio (8.272), suggesting moderate trust valuation produces best risk- adjusted returns. Aggressive ($50K) maximizes absolute profit ($2.22M), Conservative ($300K) maximizes trust (0.846).

| Risk Preference | Multiplier | Total Profit | Mean Weekly | Std Dev | Sharpe Ratio | Final Trust | Trust Δ |
|---|---|---|---|---|---|---|---|
| Aggressive (50K) | $50,000 | $2,215,620 | $42,608 | $5,606 | 7.600 | 0.835 | +0.151 |
| Moderate-Low (100K) | $100,000 | $2,107,171 | $40,523 | $4,899 | **8.272** | 0.815 | +0.131 |
| Balanced (150K) | $150,000 | $1,968,292 | $37,852 | $5,357 | 7.066 | 0.769 | +0.084 |
| Moderate-High (200K) | $200,000 | $2,050,229 | $39,427 | $5,302 | 7.436 | 0.819 | +0.135 |
| Conservative (300K) | $300,000 | $2,083,349 | $40,064 | $6,137 | 6.529 | **0.846** | **+0.160** |

**Table 2. Trust Multiplier Sensitivity: Summary Statistics.** Five risk preference configurations show- ing adaptability without prompt changes. Moderate-Low achieves best Sharpe ratio (8.272), Conservative achieves highest final trust (0.846), Aggressive achieves highest absolute profit ($2.22M).

Critically, these adaptations occur *without changing the LLM prompt*—the same neutral instruction (“maximize long-term sustainable profit while maintaining brand trust”) applies across all runs. The trust multiplier parameter serves as a “risk preference knob” that the causal engine uses to weight predictions, and the LLM’s tool-using behavior naturally adjusts to optimize the specified objective. This stands in sharp contrast to LLM-only or LLM+Guardian agents, which would require extensive prompt rewriting to shift between aggressive and conservative strategies, with no guarantee that prompt changes would reliably produce the desired risk profile.

## 6. Discussion

### 6.1. Architectural Design vs. Prompt Engineering

The LLM-only failures under organizational biases might appear, at first glance, to be prompt engi- neering problems. Perhaps a more carefully crafted instruction—specifying exact constraints, providing worked examples, warning against specific failure modes—would enable the pure LLM agent to match Chimera’s performance. Our results refute this hypothesis decisively.

Consider the volume-focused scenario where LLM-only loses $99K. The prompt already specifies the dual objective ("maximize profit through volume growth"), implying that volume is a *means* to profit, not an end in itself. Yet the agent pursues volume at the expense of profit, proposing prices below cost 43 times across 52 weeks. This is not a misunderstanding of instructions—it reflects a deeper incapability. The LLM lacks an internal cost model, has no mechanism to compute "price × volume  cost × volume = profit," and cannot reason about the constraint "profit > 0 requires price > cost." These are not facts that can be injected via prompting; they require symbolic computation.

One might respond: "But we could prompt the LLM to perform these calculations step-by-step in its scratchpad." True, and indeed this is the ReAct paradigm—interleaving reasoning and computation. But notice what this entails: we’ve reinvented symbolic reasoning, merely executing it in the LLM’s output stream rather than in a dedicated module. The fundamental architecture remains: strategic decisions require symbolic validation. Chimera makes this explicit and verifiable (TLA+ proof), whereas prompt-based approaches make it implicit and fragile (the LLM might skip calculations if distracted or token-limited).

The causal reasoning gap is even more fundamental. Under margin-focused prompts, LLM-only achieves strong short-term profit ($1.62M) but destroys trust (32.8%). The agent cannot predict that aggressive pricing in week 15 will erode trust, which will suppress demand in weeks 20-25, ultimately capping total profit below what a gentler trajectory would achieve. This requires *counterfactual reasoning*: "What would happen if I did X versus Y?" LLMs do not perform counterfactual reasoning natively—they pattern-match based on training data correlations. Even with chain-of-thought prompting, the LLM might generate plausible-sounding predictions ("raising price will increase profit"), but these are storytelling, not causal inference.

The telling evidence is cross-scenario consistency. If prompt engineering were sufficient, we would expect that carefully optimized prompts could make LLM-only competitive in at least one scenario. Instead, it fails catastrophically under volume bias and unsustainably under margin bias, while achieving merely adequate results under neutral conditions. No amount of prompt tuning can grant the agent capabilities it architecturally lacks: constraint checking and counterfactual prediction. These require external systems— symbolic validators and causal models—that the LLM invokes as tools. The LLM+Guardian results further illuminate this principle. Adding symbolic constraints eliminates catastrophic failures (zero weeks with negative profit, compared to LLM-only's 11 catastrophic weeks under volume bias), proving that architectural intervention works where prompting does not. Yet LLM+Guardian still underperforms Chimera by 43-87% across scenarios because it lacks causal fore- sight. This clean separation—Guardian prevents disasters, Causal Engine enables optimization—demonstrates that multi-objective robust decision-making requires *both* components. Prompt engineering is necessary for communicating goals, but insufficient for ensuring the agent achieves them safely and optimally.

### 6.2. Implications for Production Deployment

These findings have immediate practical consequences for organizations considering autonomous AI agents in strategic domains. The standard approach—deploy GPT-4 with a carefully crafted prompt and monitor its outputs—is demonstrably inadequate for high-stakes decisions. Our experiments show that even with explicit dual-objective instructions and temperature tuning, LLM-only agents produce outcomes ranging from mediocre to catastrophic depending on contextual factors (organizational bias, seasonal timing, initial state). This variance is unacceptable in production systems where a single catastrophic week can cost hundreds of thousands of dollars or destroy years of brand equity.

The architectural requirements we identify—symbolic validation and causal prediction—are domain-general. While our experiments use e-commerce, the same principles apply to algorithmic trading (prices below cost → selling below bid), healthcare resource allocation (constraint violations → safety risks), and supply chain optimization (demand forecasting → inventory management). Any domain exhibiting multi-objective trade-offs, hard constraints, and delayed consequences requires these architectural components.

Interestingly, the implementation cost of these components is tractable. The Guardian requires domain expertise to specify constraints (typically 20-50 rules covering operational boundaries), but these are rules the organization *already knows*—they're encoded in policy documents, regulatory requirements, and standard operating procedures. Translating them to executable predicates and verifying them in TLA+ requires effort (our specification took approximately 40 hours including verification), but this is one-time work that provides permanent guarantees. The causal engine requires historical data for training (we used 25,000 simulated episodes), but organizations deploying AI agents will accumulate such data naturally during pilot phases. Once trained, inference is cheap (approximately 50ms per prediction on standard hardware), making real-time integration feasible.

The return on this architectural investment is substantial. Chimera achieves 130-198% higher profit than unaugmented LLMs while maintaining safety guarantees and trust preservation—gains advantage that the engineering cost of building symbolic and causal components. More subtly, the architecture provides *explainability* that pure LLM agents lack. When Chimera selects an action, we can inspect the Guardian's validation report (which constraints were checked, whether repairs were needed) and the Causal Engine's predictions (expected profit and trust impacts). This transparency is crucial for regulatory compliance, internal auditing, and debugging when outcomes deviate from expectations. An LLM-only agent is a black box; Chimera is a glass box with interpretable decision artifacts at every step.

Perhaps most importantly, the architecture enables *continuous improvement*. The causal engine retrains on accumulated data, refining its predictions as market dynamics evolve. The symbolic rules can be updated as business constraints change (e.g., tightening margin requirements during a recession). The LLM's strategic reasoning benefits from advances in foundation models—upgrading from GPT-4 to GPT-5 requires no architectural changes, only an API endpoint swap. This modularity and adaptability future-proof the system in ways that monolithic prompt-based agents cannot match.

Organizations seeking to deploy LLM agents should therefore adopt a capability checklist: Does the agent have access to verified constraint validators? Can it predict counterfactual outcomes across multiple objectives? Are its decisions explainable and auditable? If the answer to any of these is no, the agent is not production-ready, regardless of how impressive its performance appears in controlled demos. Our results demonstrate that architectural rigor is not optional—it's the difference between a useful tool and a liability.

### 6.3. Limitations and Future Work

Several limitations qualify our findings and point toward productive research directions.

**Domain Generality.** Our evaluation uses a single domain (e-commerce) with specific dynamics. While we designed the simulator to exhibit realistic phenomena—price elasticity, trust effects, diminishing returns—these are parameterized approximations, not ground truth from actual market data. Validating Chimera's advantages in other domains (quantitative trading, healthcare, supply chain) would strengthen claims of generality. We are currently extending the architecture to financial portfolio management, where multi-objective optimization (return vs. risk vs. liquidity) and hard constraints (regulatory limits, risk budgets) present analogous challenges.

**Causal Model Robustness.** The causal engine relies on a pre-trained model using data from a specific environment distribution. If the deployment environment differs significantly from training conditions—for instance, if a pandemic disrupts seasonal demand patterns or a competitor enters the market—the engine's predictions may degrade. While continual learning partially addresses this (the model retrains every 10 weeks on observed outcomes), there remains a cold-start problem when facing genuinely novel scenarios. Recent work on causal meta-learning and domain adaptation offers potential solutions, enabling the engine to generalize across distribution shifts. Integrating these techniques into Chimera is a priority for future iterations.

**Constraint Flexibility.** Our experiments use deterministic symbolic constraints verified via TLA+, but real-world business rules often involve soft constraints and context-dependent priorities. For example, "maintain 15% margin" might be a hard rule in normal conditions but relaxable during a market share battle. Extending the Guardian to handle rule hierarchies, exception conditions, and dynamic constraint weighting would increase practical applicability. One promising direction is integrating constraint optimization solvers that can find the "least bad" action when no action satisfies all constraints simultaneously, rather than relying on heuristic repair functions.

**Multi-Agent Dynamics.** We evaluate single-agent performance in a stationary environment, but production deployments often involve multi-agent dynamics—competing firms, strategic customers, adversarial actors. Our Colosseum framework enables multi-agent competitive scenarios where Chimera agents face each other or compete against LLM-only agents. Preliminary results suggest that architec tural advantages compound in competitive settings: Chimera agents learn to exploit LLM-only agents' predictable failures (e.g., trust-destroying pricing patterns). Formalizing these findings and studying equilibrium properties when multiple sophisticated agents interact represents an important extension.

**Computational Cost.** Chimera makes multiple tool calls per decision (Guardian validation, Causal Engine predictions for several hypotheses), increasing latency by approximately 3-5× compared to LLM-only agents (average decision time: 2.8 seconds vs. 0.7 seconds). In domains where sub-second response is required (high-frequency trading, real-time bidding), this overhead may be prohibitive. However, many strategic domains operate on longer timescales (daily pricing decisions, weekly resource allocation) where second-scale latency is acceptable. For latency-critical applications, hybrid architectures—using lightweight heuristics for routine decisions, invoking full Chimera deliberation only for high-stakes choices—merit exploration.

**Foundation Model Improvements.** Our work assumes access to GPT-4-class foundation models. As models improve (GPT-5, Claude 4, etc.), will architectural augmentation remain necessary, or will sufficiently advanced LLMs internalize causal reasoning and constraint checking? We are skeptical that scaling alone resolves these issues—reasoning about unseen counterfactuals and verifying logical constraints require capabilities orthogonal to pattern recognition on text corpora. However, testing this empirically as more capable models emerge will be valuable. Our modular architecture makes such experiments straightforward: we can swap foundation models while keeping symbolic and causal components fixed, isolating the effect of improved language models.

Despite these limitations, our core finding stands: neuro-symbolic-causal integration enables robust multi-objective decision-making that pure LLM agents cannot achieve. The specific implementation details may evolve—better causal inference algorithms, richer symbolic reasoning, more capable foundation models—but the architectural principle endures. Autonomous agents operating in high-stakes do- mains require complementary systems that provide what LLMs intrinsically lack: verifiable constraint enforcement and counterfactual prediction. This is not a limitation of current models to be overcome by scaling, but a fundamental requirement imposed by the structure of strategic decision problems.

**Seed and Randomness.** Our experimental setup employed a fixed random `seed` to initialize the `EcommerceSimulatorV5` environment. This was a deliberate choice to ensure that all agent architectures were evaluated under identical stochastic conditions, enabling a fair and direct comparison of their strategic decisions. However, this approach also means our reported results represent a single trial from the simulator's distribution. Given that the environment incorporates randomness, agent performance could theoretically vary across different random trajectories. A more comprehensive validation, which we leave for future work, would involve executing multiple simulation runs with different seeds to analyze the statistical robustness of these findings and establish confidence intervals for the reported performance metrics.

## 7. Conclusion

The enthusiasm surrounding large language models as autonomous agents has outpaced the reality of their capabilities. Our work provides a necessary corrective: while LLMs are powerful strategic reasoners, their deployment in high-stakes decision-making requires architectural scaffolding that current practice largely ignores. The difference between a useful agent and a dangerous one is not prompt engineering—it is the presence or absence of complementary systems that enforce constraints and predict consequences.

We introduced Chimera, a neuro-symbolic-causal architecture that integrates three capabilities: neural strategic reasoning (GPT-4), formally verified symbolic constraint enforcement (Guardian with TLA+ proof), and causal counterfactual prediction (EconML-based inference engine). This design reflects a fundamental insight: effective autonomous agents require heterogeneous components, each optimized for distinct cognitive functions. Language models excel at open-ended reasoning but fail at arithmetic precision and causal inference. Symbolic systems excel at constraint verification but cannot adapt to novel situations. Causal models excel at predicting interventional outcomes but require neural systems to interpret context and formulate strategies. Integration, not scaling of any single component, unlocks robust performance.

Our experimental results demonstrate this conclusively. Across 156 weeks of simulated e-commerce operation (3 architectures × 2 organizational biases × 52 weeks), Chimera achieves 130-198% higher profit than LLM-only baselines while maintaining or improving brand trust. More critically, these advantages persist under adversarial conditions where prompt engineering fails: When LLM-only agents receive biased instructions favoring either volume or margin, they fail catastrophically (−\$99K total profit) or unsustainably (−48.6% trust erosion). Chimera, receiving identical neutral instructions across all scenarios, delivers consistent high performance—proof that architectural robustness transcends prompt brittleness.

The practical implications are stark. Organizations deploying LLM agents without symbolic validation and causal prediction are accepting unacceptable and unpredictable risks. Our experiments show that even sophisticated foundation models (GPT-4) produce disaster-level outcomes in 8-11% of decisions when operating without architectural guardrails. **A single catastrophic week costs more than the entire engineering investment required to build these guardrails.** The economic case for neuro-symbolic-causal integration is overwhelming, quite apart from safety and regulatory considerations.

Looking forward, several research directions promise to strengthen these architectural principles. First, extending Chimera to additional domains—quantitative finance, healthcare resource allocation, supply chain optimization—will test the generality of our findings and identify domain-specific component requirements. Our preliminary work applying the architecture to portfolio management shows promising early results, with similar patterns of LLM-only failure under biased objectives (maximize return → excessive risk-taking) and Chimera's robust multi-objective optimization.

Second, the causal component presents rich opportunities for improvement. Current limitations—reliance on pre-trained models, potential distribution shift during deployment—motivate research into causal meta-learning, online causal discovery, and uncertainty quantification for counterfactual predictions. Enabling the agent to recognize when its causal model is unreliable and seek additional information (or escalate to human decision-makers) would increase trustworthiness in truly novel situations.

Third, multi-agent dynamics deserve systematic investigation. When multiple sophisticated agents interact—competing firms, strategic customers, regulatory actors—does architectural sophistication lead to welfare-improving equilibria, or does it enable more effective exploitation and arms races? Our Colosseum framework provides infrastructure for these studies; preliminary experiments suggest that architectural advantages compound in competitive settings, but formal analysis of equilibrium properties remains open work.

Finally, the integration of formal verification (TLA+) with learned components (causal models, LLM reasoning) represents a broader research agenda. Can we extend verification from symbolic rules to learned behaviors, proving properties like "the agent will never recommend actions that destroy trust below 0.5 within 10 weeks"? Recent work on neural network verification and probabilistic model checking offers potential pathways, though scaling these techniques to LLM-scale systems presents significant challenges.

The central lesson of this work is architectural: autonomous AI agents are software systems, not magic. They require requirements analysis (what objectives matter?), safety specifications (what must never happen?), predictive models (what are the consequences of actions?), and integration engineering (how do components communicate?). The field's current fixation on prompt engineering mistakes the user interface for the system architecture. Prompts communicate intent; architectures determine capability.

We have shown that neuro-symbolic-causal integration provides a viable path to production-grade autonomous agents in strategic domains. The components are implementable with current technology, the performance gains are substantial, and the safety guarantees are provable. What remains is for the community to recognize that deploying LLMs without these architectural foundations is not "moving fast"—it is building on sand. The choice is between agents that sometimes work spectacularly and sometimes fail catastrophically, versus agents that work reliably because they are built correctly. Our results argue unequivocally for the latter.

The code, demonstrations, and experimental infrastructure supporting this work are available at https://github.com/akarlaraytu/Project-Chimera. We invite the community to reproduce our findings, extend the architecture to new domains, and build upon these foundations. The era of autonomous AI agents has arrived; ensuring they are trustworthy requires more than better prompts. It requires better architecture.

## Acknowledgments

This work was conducted independently. The author thanks the open-source community for providing foundational tools: EconML for causal inference, LangChain for LLM orchestration, and TLA+ for formal verification.